\documentclass{article}

\usepackage{microtype}
\usepackage{graphicx}
\usepackage{subcaption}
\usepackage{booktabs} 
\usepackage{makecell}
\usepackage{multirow}

\usepackage{hyperref}

\usepackage[preprint]{icml2026}

\usepackage{amsmath}
\usepackage{amssymb}
\usepackage{mathtools}
\usepackage{amsthm}

\usepackage[capitalize,noabbrev]{cleveref}

\theoremstyle{plain}

\theoremstyle{definition}

\theoremstyle{remark}

\usepackage[textsize=tiny]{todonotes}

\icmltitlerunning{TACLer: Tailored Curriculum Reinforcement Learning for Efficient Reasoning}

\begin{document}

\twocolumn[
  \icmltitle{TACLer: Tailored Curriculum Reinforcement \\ Learning for Efficient Reasoning}



  \icmlsetsymbol{equal}{*}

  \begin{icmlauthorlist}
    \icmlauthor{Huiyuan Lai}{rug}
    \icmlauthor{Malvina Nissim}{rug}
  \end{icmlauthorlist}

  \icmlaffiliation{rug}{Center for Language and Cognition (CLCG), University of Groningen, The Netherlands}

  \icmlcorrespondingauthor{Malvina Nissim}{m.nissim@rug.nl}

  \icmlkeywords{Machine Learning, ICML}

  \vskip 0.3in
]



\printAffiliationsAndNotice{}  

\begin{abstract}
  Large Language Models (LLMs) have shown remarkable performance on complex reasoning tasks, especially when equipped with long chain-of-thought (CoT) reasoning. However, eliciting long CoT typically requires large-scale reinforcement learning (RL) training, while often leading to overthinking with redundant intermediate steps. To improve learning and reasoning efficiency, while preserving or even enhancing performance, we propose \textbf{TACLer}, a model-tailored curriculum reinforcement learning framework that gradually increases the complexity of the data based on the model's proficiency in multi-stage RL training. TACLer features two core components: (i) tailored curriculum learning that determines what knowledge the model lacks and needs to learn in progressive stages; (ii) a hybrid \textit{Thinking}/\textit{NoThinking} reasoning paradigm that balances accuracy and efficiency by enabling or disabling the \textit{Thinking} mode. Our experiments show that TACLer yields a twofold advantage in learning and reasoning: (i) it reduces computational cost, cutting training compute by over 50\% compared to long thinking models and reducing inference token usage by over 42\% relative to the base model; and (ii) it improves accuracy by over 9\% on the base model, consistently outperforming  state-of-the-art \textit{Nothinking} and \textit{Thinking} baselines across four math datasets with complex problems.\footnote{Code and model are available at \url{https://github.com/laihuiyuan/tacler}.}
\end{abstract}

\section{Introduction}

Many recent developments in Large Language Models (LLMs) have focused on improving their ability to solve problems involving complex reasoning, which has long been considered one of the most challenging tasks~\citep{wei2022chain, wang2023selfconsistency, zhou2023leasttomost, yue2024mammoth}. Particularly, DeepSeek-R1~\citep{deepseekai2025r1} and OpenAI’s o1~\citep{openai2025o1} show that long chain-of-thought (CoT,~\citealp{wei2022chain}) sequences can be used to boost reasoning capabilities through large-scale reinforcement learning (RL), playing a crucial role in solving complex problems. The long CoT responses often contain a thinking process with reflection, backtracking, and self-validation, which helps to form the solution and arrive at the final answer. This test-time scaling paradigm~\citep{wu2024scaling, muennighoff2025s1}, enhances reasoning abilities at the cost of significantly increased token usage. Consequently, this introduces two major bottlenecks: 
(i) training models with reinforcement learning over large contexts demands substantial computational resources, e.g., a small model with 1.5 billion parameters requires 70,000 A100 GPU hours~\citep{deepscaler2025};
and (ii) so-called ``overthinking" can occur during response generation, leading to redundant reasoning steps and an even heavier computational cost~\citep{chen2025do, luo2025o1pruner, arora2025tlmre}.

To improve training efficiency, recent studies have proposed approaches such as the iterative lengthening scheme~\citep{deepscaler2025} and curriculum learning~\citep{song2025fastcurl}. Regarding overthinking, recent works explore efficient reasoning by reducing token usage in the thinking process, applying techniques such as length-based rewards~\citep{arora2025tlmre, shen2025dast}, thinking pruning~\citep{luo2025o1pruner}, and training models with adaptive thinking~\citep{zhang2025adaptthink, tu2025autothink}. Nevertheless, these approaches often entail substantial additional training while only achieving limited reductions in response length, or may even degrade overall performance. Furthermore, models that rely on adaptive reasoning strategies tend to constrain users' options since in some scenarios users might prefer to access long thinking traces even for relatively simple tasks, for purposes such as interpretability, verification, and education. We argue that a more effective paradigm should enable the coexistence of \textit{Thinking} and \textit{Nothinking} modes, balancing computational efficiency and model performance while providing flexible user control.

To this end, we propose TACLer, a tailored curriculum reinforcement learning framework, aimed at improving both learning and reasoning efficiency. First, unlike conventional curriculum learning that treats difficulty as a standardised concept, TACLer adapts the learning process to the model’s evolving proficiency. Specifically, we gradually increase the complexity of the data based on the model’s solution quality, enabling efficient learning of challenging tasks that are difficult to learn from scratch. Second, we adopt a hybrid \textit{Thinking}/\textit{NoThinking} reasoning paradigm to balance model accuracy and efficiency. Our approach allows switching between concise reasoning for higher efficiency and long thinking reasoning for improved accuracy, thus reducing unnecessary computation at the user's discretion.

We demonstrate that TACLer reduces the required training compute by over 50\% compared to other long thinking models such as DeepScaleR-1.5B-Preview~\citep{deepscaler2025}. Particularly, our extensive experimental results on four mathematical reasoning benchmarks (MATH 500, AMC, AIME 2024 and AIME 2025) show that TACLer achieves the best performance compared to various \textit{Nothinking} and \textit{Thinking} baselines, while significantly reducing the average number of token usage by over 42\%, thereby effectively mitigating the overthinking problem.

In summary, the key contributions of our paper are as follows: (1) we introduce tailored curriculum learning, which determines what knowledge the model specifically lacks and needs to learn in progressive stages; (2) we incorporate both  \textit{Thinking} and \textit{NoThinking} modes, showing that this can effectively balance reasoning accuracy and efficiency; (3) we propose TACLer, a novel RL-based framework that enables language models to learn more efficiently through tailored curriculum learning and to reason effectively via hybrid reasoning modes; and (4) we report extensive experiments and analysis to validate the efficiency of TACLer on both learning and reasoning.

\section{Preliminaries: Learning and Reasoning Efficiency}
\label{sec:preliminary}

\begin{figure}[t]
  \centering
  \includegraphics[width=0.42\textwidth]{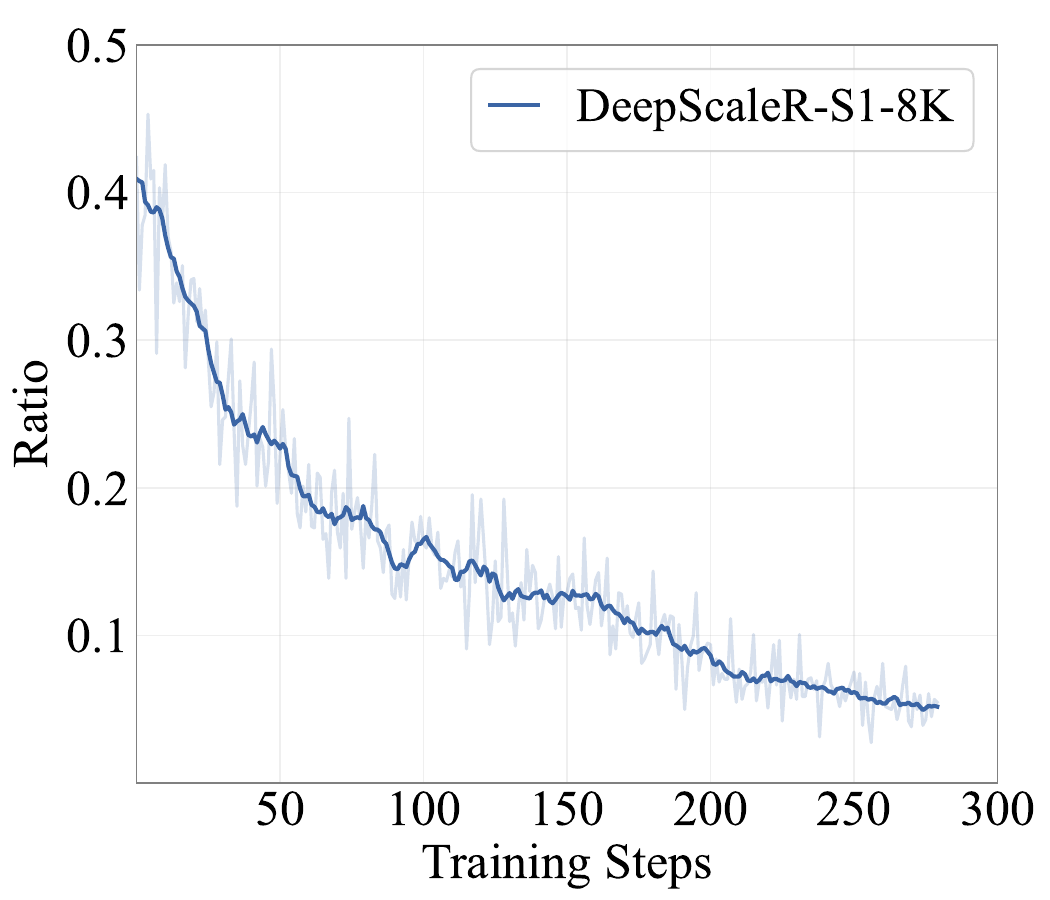}
  \caption{Response clipping ratio in the first training stage of DeepScaleR-1.5B-Preview.}
  \label{fig:deepscaler-clip}
\end{figure}

The computational burden of training language models is often enormous, especially for tasks involving long thinking reasoning that require large context windows. For instance, \citet{deepscaler2025} indicates that replicating DeepSeek-R1 experiments with $\geq$32K context and $\sim$8000 training steps would take at least 70,000 A100 GPU hours, even for a small model with 1.5B parameters. To mitigate this problem, \citet{deepscaler2025} introduce an iterative lengthening scheme, which progressively increases the context length from 8K to 24K, thereby guiding the model to utilise the context more efficiently. However, as shown in Figure~\ref{fig:deepscaler-clip}, more than 40\% of the responses generated in the first training stage are still truncated, revealing a substantial inefficiency in the training process. Although \citet{song2025fastcurl} propose a curriculum learning framework that gradually increases the context length based on input length with the aim of facilitating progressive learning from easy to hard, almost 35\% of responses remain truncated in the first stage. The primary reason is that this is an arbitrary method for curriculum learning that does not take into account the model's mastery of specific problems.

\begin{figure*}[h]
    \centering
    \begin{subfigure}[t]{0.45\linewidth}
        \centering
        \includegraphics[width=\linewidth]{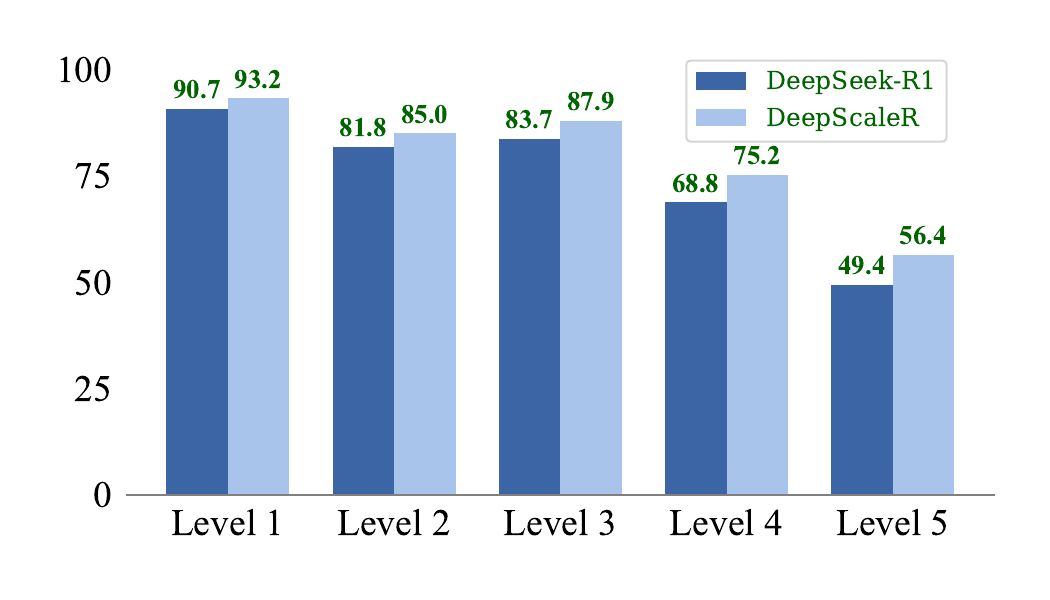}
        \caption{Reasoning accuracy (\%).}
        \label{fig:pre-acc}
    \end{subfigure}
    \begin{subfigure}[t]{0.45\linewidth}
        \centering
        \includegraphics[width=\linewidth]{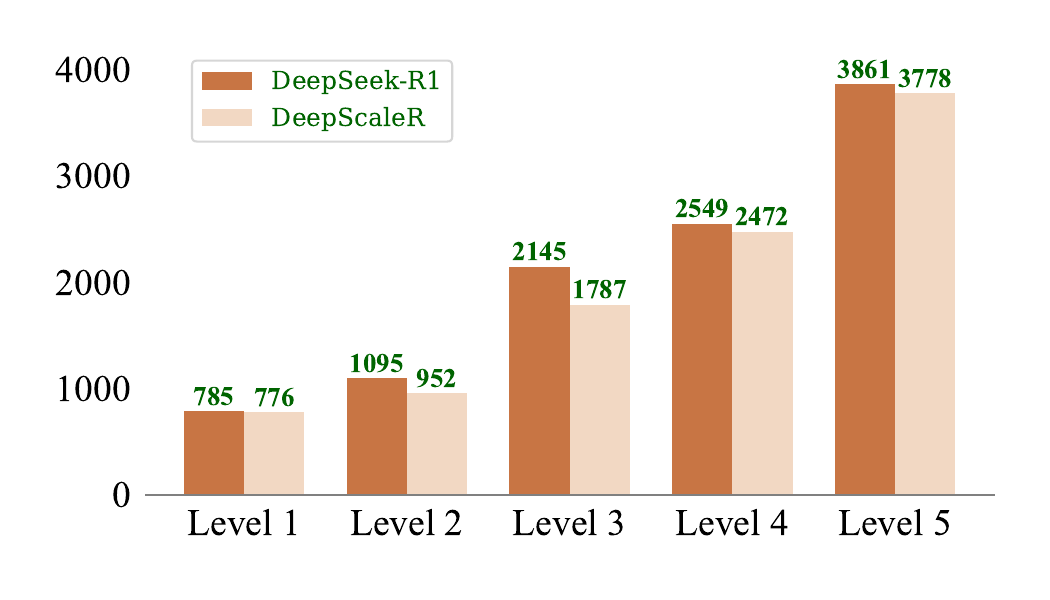}
        \caption{Response length.}
        \label{fig:pre-len}
    \end{subfigure}
    \caption{Comparison of DeepSeek-R1-Distill-Qwen-1.5B and DeepScaleR-1.5B-Preview using \textit{NoThinking} mode across different difficulty levels of MATH500 dataset.}
    \label{fig:pre-acc-len}
\end{figure*}

Regarding reasoning efficiency, Figure~\ref{fig:pre-acc-len} presents the comparison of DeepSeek-R1 and DeepScaleR in the \textit{NoThinking} mode (see Sec.~\ref{subsec:hybrid} for details), where the latter is trained based on the former to enhance the long CoT reasoning capability in the \textit{Thinking} mode. Interestingly, DeepScaleR in the \textit{NoThinking} mode exhibits consistent performance gains across questions of varying difficulty, accompanied by shorter responses. This shows that \textit{Thinking} and \textit{NoThinking} modes can coexist in the model and can be enhanced through RL training. Based on the above observation, a natural question arises: Can LLMs learn efficiently through a better curriculum learning strategy while leveraging a hybrid mode for efficient reasoning? In this work, we introduce TACLer, a novel RL-based framework that targets math reasoning tasks, which we detail in the next section.

\begin{figure*}[t]
  \centering
  \includegraphics[width=0.9\textwidth]{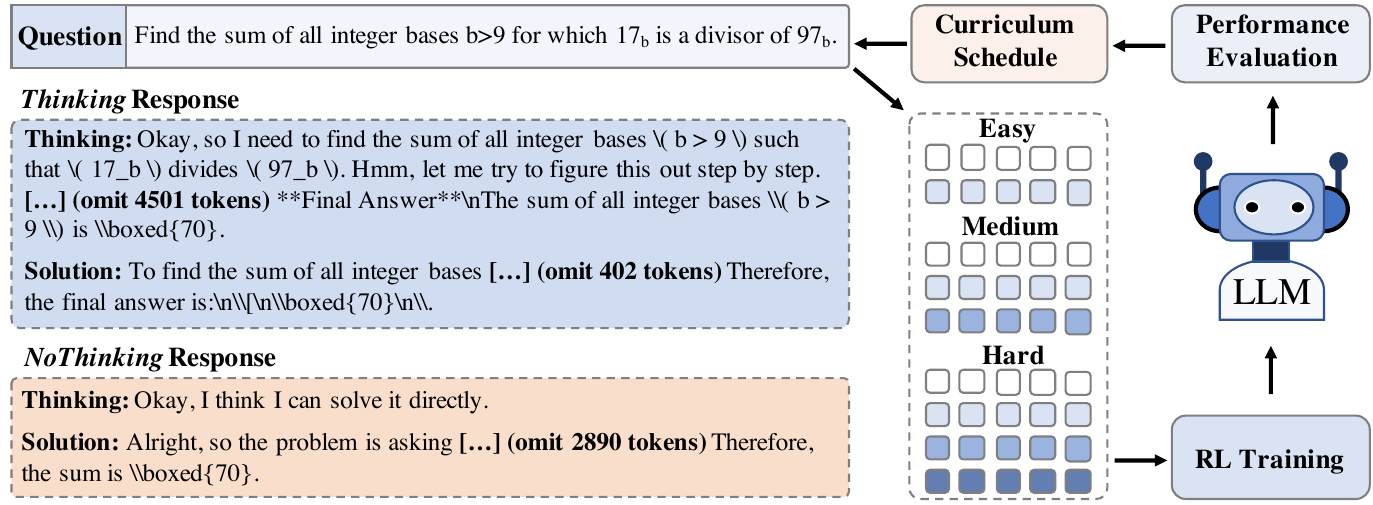}
  \caption{Overview of TACLer, our tailored curriculum reinforcement learning framework.}
  \label{fig:framework}
\end{figure*}

\section{Methodology}

Figure~\ref{fig:framework} gives an overview of  
TACLer and its three components: (1) performance evaluation and curriculum schedule; (2) hybrid reasoning mode; and (3) RL training. Measuring performance is needed to create the curriculum schedule, with the goal of identifying the model's knowledge gap and enabling efficient learning across progressive RL stages. The hybrid reasoning mode, consisting of \textit{Thinking} and \textit{NoThinking}, achieves more efficient reasoning by balancing performance and computational cost during inference.


\subsection{Tailored Curriculum Learning}
\label{sub:tailored}

As shown in Section~\ref{sec:preliminary}, the imbalanced data distribution poses a significant challenge: during early training, a substantial portion of samples are overly difficult. The resulting truncation not only leads to wasted computation but also hinders effective learning, thus substantially reducing training efficiency. To address this issue, we employ curriculum learning \citep{bengio2009curriculum}, which gradually increases the data complexity throughout the learning process. 

Our method leverages the notion of training data difficulty, where difficulty is defined by the model’s ability to solve the corresponding problems. 
We perform inference for all training data with greedy decoding and a context size of 8k, and categorise instances based on the responses and answers: (1) instances with the correct final answer; (2) instances with a full reasoning process within the maximum length but an incorrect final answer; and (3) instances where the model fails to generate a full solution within the maximum length. Based on these results, we can quantify problem-specific difficulty in terms of the model’s proficiency. In turn, this enables us to organise training samples according to a tailored curriculum learning strategy, thereby allowing each model to tackle problems progressively from those it finds easier to solve to the harder ones.

During training, we merge samples from the first and second groups in the above inference results, ensuring that the training data is neither dominated by overly difficult nor overly simple problems, to implement a combined \textit{review and learning} paradigm. This strategy enables the model to reinforce previously acquired knowledge while learning new skills, thus mitigating forgetting and enhancing overall learning efficiency. This entire process---inference, difficulty categorisation, curriculum learning---is repeated twice in our framework.\footnote{See Appendix~\ref{app:setting} for details on computational cost.} In the third stage, we train on the full dataset using the model updated through the first two iterations to consolidate the learning results.

\subsection{Hybrid Reasoning Mode}
\label{subsec:hybrid}

Based on our preliminary findings in Section~\ref{sec:preliminary}, we adopt a hybrid reasoning mode that integrates both \textit{Thinking} and \textit{NoThinking}. Generally, many existing reasoning models, such as DeepSeek-R1, adopt the response structure that consists of a long thinking process marked by $<$\texttt{thinking}$>$ and $<$/\texttt{thinking}$>$, followed by the final solution. Our hybrid reasoning mode is defined as follows.

\paragraph{Thinking} This refers to the default structure for querying language models, in which the model is prompted to generate the detailed thinking process within given special markers. Based on this, the model then derives a solution and generates a final answer presented as its final prediction. This approach encourages the model to reflect, backtrack, and self-verify during the reasoning process. 

\paragraph{NoThinking} This is a concise response structure that includes the final solution and answer without the detailed thinking process within special markers. Instead of using the prompt ``\texttt{Okay, I think I have finished thinking}'' to induce the model to state that it has completed its reasoning~\citep{ma2025reasoning}, we explicitly prompt the model to solve the problem without thinking: $<$\texttt{thinking}$>$ \texttt{Okay, I think I can solve it directly.} $<$\texttt{/thinking}$>$.

We train TACLer in the hybrid reasoning mode across all three stages, arguing that the two modes can be mutually beneficial. Specifically, a model trained solely in the \textit{NoThinking} mode may compensate for missing explicit thinking by expanding its reasoning steps, reducing the risk of producing incorrect answers. In other words, \textit{NoThinking} does not simply mean ``short'' (see Sec.~\ref{sub:analysis}.) In contrast, when trained jointly with the \textit{Thinking} mode, the model can leverage a compression effect, where improved reasoning ability is distilled into shorter responses. This integration has the potential to yield concise responses without sacrificing accuracy; in fact, it may even improve it.

\subsection{Training with GRPO}

To train our model, we adopt the Group Relative Policy Optimisation (GRPO,~\citealp{shao2024deepseek}) along with several improvements recently proposed by the research community. For each problem $q$, the algorithm samples a group of responses \{$o_1, o_2, ... , o_G$\} from the old policy $\pi_{old}$, and updates the policy $\pi_{\theta}$ by maximising the following objective:
\begin{align}
\label{eq:grpo}
\mathcal{J}_{\text{GRPO}}(\theta) = &
\mathbb{E}_{q \sim \mathcal{D}, \{o_i\}_{i=1}^G \sim \pi_{\theta_{\text{old}}}(\cdot \mid q)} \\
& \frac{1}{G} 
\sum_{i=1}^G
\min \Bigg(
\frac{\pi_\theta(o_i \mid q)}{\pi_{\theta_\text{old}}(o_i \mid q)} A_{i},\nonumber\\ 
& \text{clip} \Big(\frac{\pi_\theta(o_i \mid q)}{\pi_{\theta_\text{old}}(o_i \mid q)}, 1 - \epsilon_{\text{low}}, 1 + \epsilon_{\text{high}} \Big) A_{i}
\Bigg) \nonumber
\end{align}

where $\epsilon_{\text{low}}$ and $\epsilon_{\text{high}}$ are hyperparameters for the clipping range of the importance sampling ratio. $A_{i}$ is the advantage, caculated by the rewards \{$r_1, r_2, ..., r_G$\} of responses in the same group:
\begin{align}
\label{eq:adva}
A_{i} = \frac{r_i-\text{mean}(\{{r_i}\}_{i=1}^{G})}{\text{std}(\{{r_i}\}_{i=1}^{G})}
\end{align}

Following prior work~\citep{deepseekai2025r1}, we employ a rule-based reward scheme, assigning a binary reward of 1 to correct final answers and 0 otherwise. We incorporate several enhancements from recent research into original GRPO, including (1) removing the KL loss to effectively unlock the full potential of the policy model without affecting training stability~\citep{hu2025open}; and (2) increasing the upper clip bounds $\epsilon_{\text{high}}$ in Eq. ~\ref{eq:grpo} while fixing the lower clip bounds $\epsilon_{\text{low}}$ to mitigate the entropy convergence problem and encourage the policy to explore diverse solutions~\citep{yu2025dapo}.

\section{Experiments}

\subsection{Experimental Setup}
\label{subsec:seup}

\paragraph{Model and Dataset} 
We select DeepSeek-R1-Distill-Qwen-1.5B (R1-Qwen, \citealp{deepseekai2025r1}) as our backbone, given its proven strong performance on mathematical tasks. For training data, we use DeepScaleR-Dataset~\citep{deepscaler2025}, containing around 40k math problems from various competitions such as AIME 1983-2023. 
The evaluation is performed on four challenging math benchmarks: MATH500~\citep{lightman2024lets}, AMC, AIME 2024, and AIME 2025, which are widely adopted for assessing LLM reasoning~\citep{deepseekai2025r1, openai2025o1}.\footnote{See Appendix~\ref{app:setting} for more details on training and evaluation.}

\begin{table*}[h]
\caption{\label{tab:thinking}
Comparison of model performance (accuracy and response length) in the long thinking mode across four mathematical reasoning datasets. Note that bold numbers indicate the best result for each dataset among different models.
}
\centering
\footnotesize
\setlength{\tabcolsep}{4pt}
\begin{tabular}{lcccccccccc}
\toprule
   & \multicolumn{2}{c}{MATH500} & \multicolumn{2}{c}{AIME 2024} & \multicolumn{2}{c}{AMC} & \multicolumn{2}{c}{AIME 2025}  & \multicolumn{2}{c}{Average}\\
   \cmidrule(lr){2-3} \cmidrule(lr){4-5} \cmidrule(lr){6-7} \cmidrule(lr){8-9} \cmidrule(lr){10-11}
   & ACC & Length $\downarrow$ & ACC & Length $\downarrow$ & ACC & Length $\downarrow$ & ACC & Length $\downarrow$ & $\Delta$ACC & $\Delta$Length $\downarrow$ \\
\hline
{\color[HTML]{7F7F7F} $\text{R1-Qwen}_{\textit{Thinking}}$} & {\color[HTML]{7F7F7F}81.2} & {\color[HTML]{7F7F7F}4856} & {\color[HTML]{7F7F7F}27.7} & {\color[HTML]{7F7F7F}12306} & {\color[HTML]{7F7F7F}60.8} & {\color[HTML]{7F7F7F}8754} & {\color[HTML]{7F7F7F}21.5} & {\color[HTML]{7F7F7F}12182} & {\color[HTML]{7F7F7F}-} & {\color[HTML]{7F7F7F}-} \\
STILL-3 & 83.6 & 3797 & 30.4 & 10605 & 66.3 & 7091 & 24.4 & 10415 & +3.4 & -17.3\%\\
DeepScaleR & 87.8 & 3030 & 40.4 & 8565 & 73.8 & 5616 & \textbf{31.3} & 8239 & +10.5 & -34.1\%\\
FastCuRL   & 87.8 & 3894 & 39.8 & 10091 & 73.9 & 6756 & 27.9 & 9723 & +9.6 & -20.2\%\\
$\text{TACLer}_\textit{Thinking}$ & \textbf{88.4} & \textbf{3010} & \textbf{42.1} & \textbf{6868} & \textbf{74.6} & \textbf{4871} & 30.8 & \textbf{6807} & \textbf{+11.2} & \textbf{-42.7\%} \\
\bottomrule
\end{tabular}
\end{table*}

\begin{table}[!t]
\caption{\label{tab:nothinking}
Comparison of model performance in the efficient thinking mode across four mathematical reasoning benchmarks. For MATH500 and AIME 2024, results for all baselines except AutoThink are taken from~\citet{zhang2025adaptthink}. 
Bold numbers indicate the best result for each dataset among the different models.
}
\centering
\footnotesize
\setlength{\tabcolsep}{0.4pt}
\begin{tabular}{lcccccc}
\toprule
   & \multicolumn{2}{c}{MATH500} & \multicolumn{2}{c}{AIME 2024}  & \multicolumn{2}{c}{Average}\\
   \cmidrule(lr){2-3} \cmidrule(lr){4-5} \cmidrule(lr){6-7}
   & ACC & Length & ACC & Length  & $\Delta$ACC & $\Delta$Length \\
\hline
{\color[HTML]{7F7F7F} $\text{R1-Qwen}_{\textit{Thinking}}$} & {\color[HTML]{7F7F7F}81.2} & {\color[HTML]{7F7F7F}4856} & {\color[HTML]{7F7F7F}27.7} & {\color[HTML]{7F7F7F}12306} & {\color[HTML]{7F7F7F}-} & {\color[HTML]{7F7F7F}-} \\
{\color[HTML]{7F7F7F} $\text{R1-Qwen}_{\textit{NoThinking}}$} & {\color[HTML]{7F7F7F}67.8} & {\color[HTML]{7F7F7F}1069} & {\color[HTML]{7F7F7F}14.8} & {\color[HTML]{7F7F7F}4689} & {\color[HTML]{7F7F7F}-13.2} & {\color[HTML]{7F7F7F}-69.9\%}\\
OverThink     & 81.2 & 4131 & 28.3 & 11269 & +0.3 & -11.7\%\\
DAST          & 83.0 & 2428 & 26.9 & 7745 & +0.5 & -43.5\%\\
O1-Pruner     & 82.2 & 3212 & 28.9 & 10361 & +1.1 & -24.8\% \\
TLMRE         & 85.0 & 3007 & 29.2 & 8982 & +2.6 & -32.5\%\\
ModelMerging  & 63.0 & 2723 & 18.1 & 10337 & -13.9 & -30.0\%\\
AdaptThink    & 86.0 &  2511 & 34.8 & 9279 & +5.9 & -36.4\%\\
AutoThink                      & 83.8 & \textbf{2128} & 31.7 & 8167 & +3.3 & -44.9\%\\
$\text{TACLer}_\textit{NoThinking}$ & \textbf{88.2} & 2532 & \textbf{39.6} & \textbf{6056} & \textbf{+9.5} & \textbf{-49.3\%}\\
\midrule
 & \multicolumn{2}{c}{AMC} & \multicolumn{2}{c}{AIME 2025}  & \multicolumn{2}{c}{Average}\\
 \cmidrule(lr){2-3} \cmidrule(lr){4-5} \cmidrule(lr){6-7}
 & ACC & Length $\downarrow$ & ACC & Length $\downarrow$ & $\Delta$ACC & $\Delta$Length $\downarrow$ \\
 \hline
{\color[HTML]{7F7F7F} $\text{R1-Qwen}_{\textit{Thinking}}$} & {\color[HTML]{7F7F7F}60.8} & {\color[HTML]{7F7F7F}8754} & {\color[HTML]{7F7F7F}21.5} & {\color[HTML]{7F7F7F}12182} & {\color[HTML]{7F7F7F}-} & {\color[HTML]{7F7F7F}-} \\
{\color[HTML]{7F7F7F} $\text{R1-Qwen}_{\textit{NoThinking}}$} & {\color[HTML]{7F7F7F}48.6} & {\color[HTML]{7F7F7F}2264} & {\color[HTML]{7F7F7F}13.3} & {\color[HTML]{7F7F7F}4062} & {\color[HTML]{7F7F7F}-10.2} & {\color[HTML]{7F7F7F}-70.4\%}\\
AdaptThink    & 67.4 & 5489 & 25.6 & 9117 & +5.4 & -31.2\%\\
AutoThink                      & 66.7 & 4596 & 23.8 & 7647 & +4.1 & -42.4\%\\
$\text{TACLer}_\textit{NoThinking}$ & \textbf{72.7} & \textbf{4312} & \textbf{27.9} & \textbf{5710} & \textbf{+9.2} & \textbf{-51.9\%}\\
\bottomrule
\end{tabular}
\end{table}

\paragraph{Baselines}
We compare TACLer with several state-of-the-art 1.5B baselines trained for long CoT reasoning: (1) \textbf{STILL-3}~\citep{chen2025an}, (2) \textbf{DeepScaleR}~\citep{deepscaler2025}, and (3) \textbf{FastCuRL}~\citep{song2025fastcurl}.
We also compare our model with state-of-the-art 1.5B baselines on efficient math reasoning: (1) \textbf{OverThink}~\citep{chen2025do}, (2) \textbf{DAST}~\citep{shen2025dast}, (3) \textbf{O1-Pruner}~\citep{luo2025o1pruner}, (4) \textbf{TLMRE}, (5) {ModelMerging}~\citep{wu2025unlock}, (6) \textbf{AdaptThink}~\citep{zhang2025adaptthink}, and (7) \textbf{AutoThink}~\citep{tu2025autothink}. All models use the DeepScaleR dataset, where methods (1)-(5) are reproduced by ~\citet{zhang2025adaptthink}, thereby providing a fair comparison.\footnote{See Appendix~\ref{app:baselines} for more details on baselines.}

\subsection{Main Results}

\paragraph{Reasoning Accuracy and Efficiency in Thinking Mode}
Table~\ref{tab:thinking} compares TACLer in\textit{Thinking} mode against several long thinking models . Across four datasets, TACLer achieves the highest accuracy on three of them, and performs slightly lower than DeepScaleR but higher than other models on AIME 2025. Overall, it achieves the \textbf{highest average accuracy}, with an average +11.2\% improvement over the base model R1-Qwen. In addition, \textbf{TACLer reduces reasoning length by 42.7\%, substantially outperforming all baselines} (e.g., -34.1\% for DeepScaleR and -20.2\% for FastCuRL), suggesting that the hybrid mode can leverage the compression effect to refine reasoning into concise responses. These results highlight that \textbf{TACLer in \textit{Thinking} mode provides a more effective balance between reasoning accuracy and response length than existing methods}.

\paragraph{Reasoning Accuracy and Efficiency in Saving Mode}
Table~\ref{tab:nothinking} compares TACLer in \textit{NoThinking} mode  with recent efficient reasoning approaches. The first observation is that TACLer consistently achieves the best overall performance, with the only exception being response length on the MATH500 dataset. Specifically, R1-Qwen initially performs poorly in \textit{NoThinking} mode, and during training, it tends to compensate for the lack of explicit reasoning by slightly increasing its token usage. After training, TACLer in \textit{NoThinking} mode substantially outperforms R1-Qwen in \textit{Thinking} mode, achieving over 9\% higher accuracy while using roughly half the tokens (49.3\% and 51.9\%).\footnote{See appendix~\ref{app:case} for example responses.} In comparison, efficient reasoning baselines such as AutoThink improve efficiency by reducing token usage, but offer smaller accuracy gains; ModelMerging even suffers from substantial performance degradation. \textbf{On average, across the four datasets, TACLer achieves higher reasoning accuracy and generates shorter responses}.

\begin{table*}[h]
\caption{\label{tab:purenothinking}
Comparison of models trained with hybrid reasoning (\textit{NoThinking} + \textit{Thinking}) versus pure \textit{NoThinking} training after 280 steps of first-stage training. Bold numbers indicate the best result for each dataset among different models.
}
\centering
\footnotesize
\setlength{\tabcolsep}{4pt}
\begin{tabular}{lcccccccccc}
\toprule
   & \multicolumn{2}{c}{MATH500} & \multicolumn{2}{c}{AIME 2024} & \multicolumn{2}{c}{AMC} & \multicolumn{2}{c}{AIME 2025}  & \multicolumn{2}{c}{Average}\\
   \cmidrule(lr){2-3} \cmidrule(lr){4-5} \cmidrule(lr){6-7} \cmidrule(lr){8-9} \cmidrule(lr){10-11}
   & ACC & Length $\downarrow$ & ACC & Length $\downarrow$ & ACC & Length $\downarrow$ & ACC & Length $\downarrow$ & $\Delta$ACC & $\Delta$Length $\downarrow$ \\
\hline
{\color[HTML]{7F7F7F} $\text{R1-Qwen}_{\textit{Thinking}}$} & {\color[HTML]{7F7F7F}81.2} & {\color[HTML]{7F7F7F}4856} & {\color[HTML]{7F7F7F}27.7} & {\color[HTML]{7F7F7F}12306} & {\color[HTML]{7F7F7F}60.8} & {\color[HTML]{7F7F7F}8754} & {\color[HTML]{7F7F7F}21.5} & {\color[HTML]{7F7F7F}12182} & {\color[HTML]{7F7F7F}-} & {\color[HTML]{7F7F7F}-} \\
Pure-NoThinking & 84.0 & 2325 & \textbf{32.1} & 8380 & 65.1 & 5223 & 24.8 & 8220 & +3.7 & -39.2\%\\
$\text{TACLer}_\textit{NoThinking}$ & 82.0 & \textbf{1472} & 25.6 & \textbf{5882} & 62.1 & \textbf{3479} & 22.5 & \textbf{5455} & +0.3 & \textbf{-59.3\%} \\
$\text{TACLer}_\textit{Thinking}$ & \textbf{84.8} & 3287 & 30.0 & 9168 & \textbf{68.2} & 5875 & 25.2 & 8873 & \textbf{+4.3} & -29.5\%\\
\bottomrule
\end{tabular}
\end{table*}

\begin{table*}[!t]
\caption{\label{tab:ablation}
Comparison of curriculum learning and direct training (training from the beginning with all data at once). Bold numbers indicate the best result for each dataset among different models.
}
\centering
\footnotesize
\setlength{\tabcolsep}{4pt}
\begin{tabular}{lcccccccccc}
\toprule
   & \multicolumn{2}{c}{MATH500} & \multicolumn{2}{c}{AIME 2024} & \multicolumn{2}{c}{AMC} & \multicolumn{2}{c}{AIME 2025}  & \multicolumn{2}{c}{Average}\\
   \cmidrule(lr){2-3} \cmidrule(lr){4-5} \cmidrule(lr){6-7} \cmidrule(lr){8-9} \cmidrule(lr){10-11}
   & ACC & Length $\downarrow$ & ACC & Length $\downarrow$ & ACC & Length $\downarrow$ & ACC & Length $\downarrow$ & $\Delta$ACC & $\Delta$Length $\downarrow$ \\
\hline
{\color[HTML]{7F7F7F} $\text{R1-Qwen}_{\textit{Thinking}}$} & {\color[HTML]{7F7F7F}81.2} & {\color[HTML]{7F7F7F}4856} & {\color[HTML]{7F7F7F}27.7} & {\color[HTML]{7F7F7F}12306} & {\color[HTML]{7F7F7F}60.8} & {\color[HTML]{7F7F7F}8754} & {\color[HTML]{7F7F7F}21.5} & {\color[HTML]{7F7F7F}12182} & {\color[HTML]{7F7F7F}-} & {\color[HTML]{7F7F7F}-} \\
\hline
\multicolumn{11}{l}{\textit{NoThinking} Mode}\\
$\text{Direct-Train}$ & 87.8 & \textbf{2217} & 38.1 & \textbf{5974} & 72.1 & \textbf{4093} & 26.9 & \textbf{5550} & +8.4 & \textbf{-53.4\%}\\
$\text{TACLer}$ & \textbf{88.2} & 2532 & \textbf{39.6} & 6056 & \textbf{72.7} & 4312 & \textbf{27.9} & 5710 & \textbf{+9.3} & -50.6\% \\
\hline
\multicolumn{11}{l}{\textit{Thinking} Mode}\\
$\text{Direct-Train}$   & 88.0 & \textbf{2962} & 37.5 & 7085 & 73.0 & 4920 & 27.5 & \textbf{6736} & +8.7 & -42.5\%\\
$\text{TACLer}$ & \textbf{88.4} & 3010 & \textbf{42.1} & \textbf{6868} & \textbf{74.6} & \textbf{4871} & \textbf{30.8} & 6807 & \textbf{+11.2} & \textbf{-42.7\%} \\
\bottomrule
\end{tabular}
\end{table*}

\subsection{Analysis and Discussion}
\label{sub:analysis}

\paragraph{Hybrid Mode Yields Concise Reasoning}
We train a model in the first stage using only \textit{NoThinking} mode and compare it with TACLer trained in the same stage, as shown in Table~\ref{tab:purenothinking}. TACLer achieves the highest accuracy in the \textit{Thinking} mode, while generating the shortest responses in the \textit{NoThinking} mode. In contrast, Pure-NoThinking achieves higher accuracy than TACLer in the \textit{NoThinking} mode but produces substantially longer responses, failing to support concise reasoning. These results show that the hybrid reasoning mode allows the model to benefit from both strategies, making reasoning under \textit{NoThinking} more concise with no heavy loss in performance.

\paragraph{Curriculum Learning Yields More Efficient and Accurate Reasoning}
We compare TACLer with Direct-Train, a model which is trained directly using all of the data at once rather than staggered based on difficulty, with all other settings staying the same (Table~\ref{tab:ablation}). While Direct-Train tends to generate shorter responses than TACLer under the \textit{NoThinking} mode, the trend is reversed in \textit{Thinking} mode, with TACLer producing more concise outputs compared to Direct-Train. Moreover, \textbf{TACLer consistently achieves a better performance across both modes in comparison to the model trained without staggered data}, showing an effective balance between conciseness and accuracy.

\begin{table}[!t]
\caption{\label{tab:upper-bound}
Comparison of TACLer to the Oracle Union of two reasoning modes. Avg. denotes the average ratio of TACLer's accuracy to the Oracle Union accuracy across all four datasets.
}
\centering
\footnotesize
\setlength{\tabcolsep}{1.5pt}
\begin{tabular}{lccccc}
\toprule
 & MATH500 & AIME2024 & AMC & AIME2024 & Avg.\\
\hline
Upper Bound	& 91.4 & 52.1 & 80.8 & 33.8 & - \\
\textit{NoThinking} & 88.2 & 39.6 & 72.7 & 27.9 & 86.3\\
\textit{Thinking} & 88.4 & 42.1 & 74.6 & 30.8 & 90.2 \\
\bottomrule
\end{tabular}
\end{table}

\paragraph{Theoretical Performance Ceiling}
We introduce an ``oracle union'' as a reference to estimate the theoretical performance ceiling of our hybrid reasoning paradigm, which selects the correct prediction from either of the two reasoning modes (\textit{Thinking}/\textit{NoThinking}). As shown in Table~\ref{tab:upper-bound}, this oracle union (Upper Bound) represents the best achievable accuracy if we could perfectly combine the strengths of both modes. Across four datasets, TACLer achieves over 86\% of this upper bound on average. 

\begin{figure*}[h]
    \centering
    \begin{subfigure}[t]{0.45\linewidth}
        \centering
        \includegraphics[width=\linewidth]{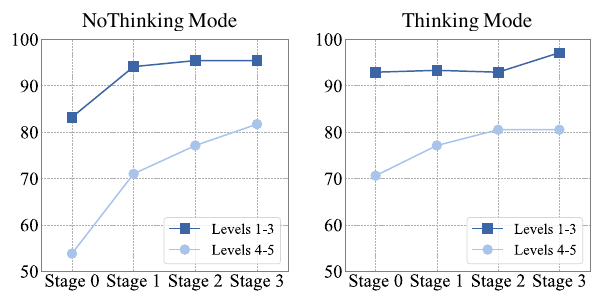}
        \caption{Reasoning accuracy.}
        \label{fig:tacl-acc}
    \end{subfigure}
    \hspace{0.02\textwidth}
    \begin{subfigure}[t]{0.45\linewidth}
        \centering
        \includegraphics[width=\linewidth]{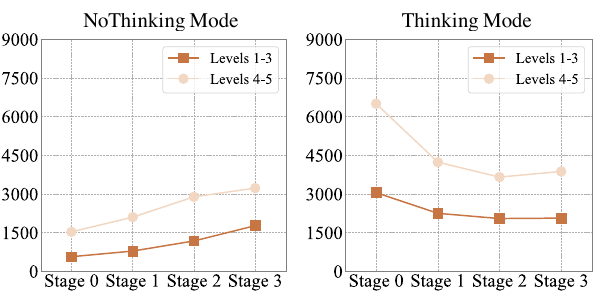}
        \caption{Response length.}
        \label{fig:tacl-len}
    \end{subfigure}
    \caption{Comparison of different training stages on different difficulty levels of MATH500. Note that state 0 represents the base model.}
    \label{fig:tacl-acc-len}
    \vspace{-0.8em}
\end{figure*}

\paragraph{TACLer's Performance at Different Stages}
Figure~\ref{fig:tacl-acc-len} shows reasoning accuracy and response length of TACLer across different training stages for two difficulty levels in MATH500 (the original 5 levels are merged into levels 1–3 and levels 4–5). For reasoning accuracy, we observe a consistent upward trend across the three training stages (see Sec.~\ref{sub:tailored}.) The largest improvement occurs in the first stage, with higher gains in the \textit{NoThinking} mode. 
Notably, in the third stage TACLer in \textit{Thinking} mode obtains higher scores on easier problems. We hypothesise that some questions categorised as “easy” by humans may actually be more challenging for the model, and can be successfully solved after the model's capabilities are substantially enhanced through curriculum learning (see examples in Appendix~\ref{app:case}.) When looking at response length, TACLer shows a different trend for \textit{Nothinking} and \textit{Thinking} modes: in the former the length of responses gradually increases with training, while in the latter it gradually decreases. This is consistent with our hypothesis: \textit{NoThinking} mode tends to compensate for the lack of explicit reasoning by slightly increasing its token usage, while \textit{Thinking} mode leverages the compression effect to refine reasoning into concise responses.

\begin{table}[!t]
\caption{\label{tab:training-info}
Training configuration comparison.
}
\centering
\footnotesize
\setlength{\tabcolsep}{4pt}
\begin{tabular}{lrrrrr}
\toprule
Stage & \makecell[c]{Context} & \makecell[c]{Batch Size} & Step & Rollout\\
\hline
\multicolumn{5}{l}{DeepScaleR-1.5B-Preview}\\
Stage 1 & 8K & 128 & 1040 & 8\\
Stage 2 & 16K & 128 & 500 & 16\\
Stage 3 & 24K & 128 & 210 & 16\\
\midrule
\multicolumn{5}{l}{FastCuRL-1.5B-Preview}\\
Stage 1 & 8K & 128 & 160 & 8\\
Stage 2 & 16K & 64 & 590 & 8\\
Stage 3 & 24K & 64 & 230 & 8\\
Stage 4 & 16K & 64 & 580 & 16\\
\midrule
\multicolumn{5}{l}{TACLer}\\
Stage 1 & 8K & 128 & 280 & 8\\
Stage 2 & 8K & 128 & 350 & 8\\
Stage 3 & 8K & 128 & 1250 & 8\\
\bottomrule
\end{tabular}
\end{table}

\paragraph{Computation Cost and Learning Efficiency}

\begin{figure*}[!t]
    \centering
    \begin{subfigure}[t]{0.31\linewidth}
        \centering
        \includegraphics[width=\linewidth]{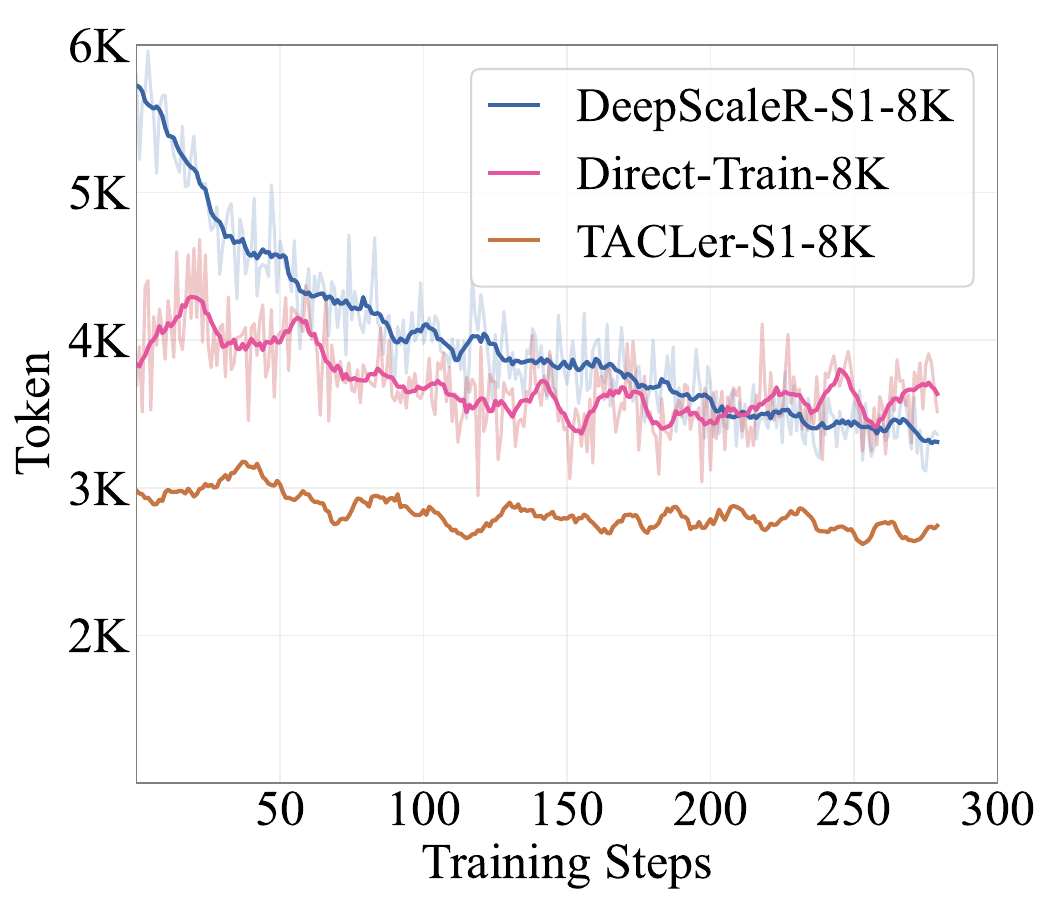}
        \caption{Response length.}
        \label{fig:s1-clip}
    \end{subfigure}
    \begin{subfigure}[t]{0.31\linewidth}
        \centering
        \includegraphics[width=\linewidth]{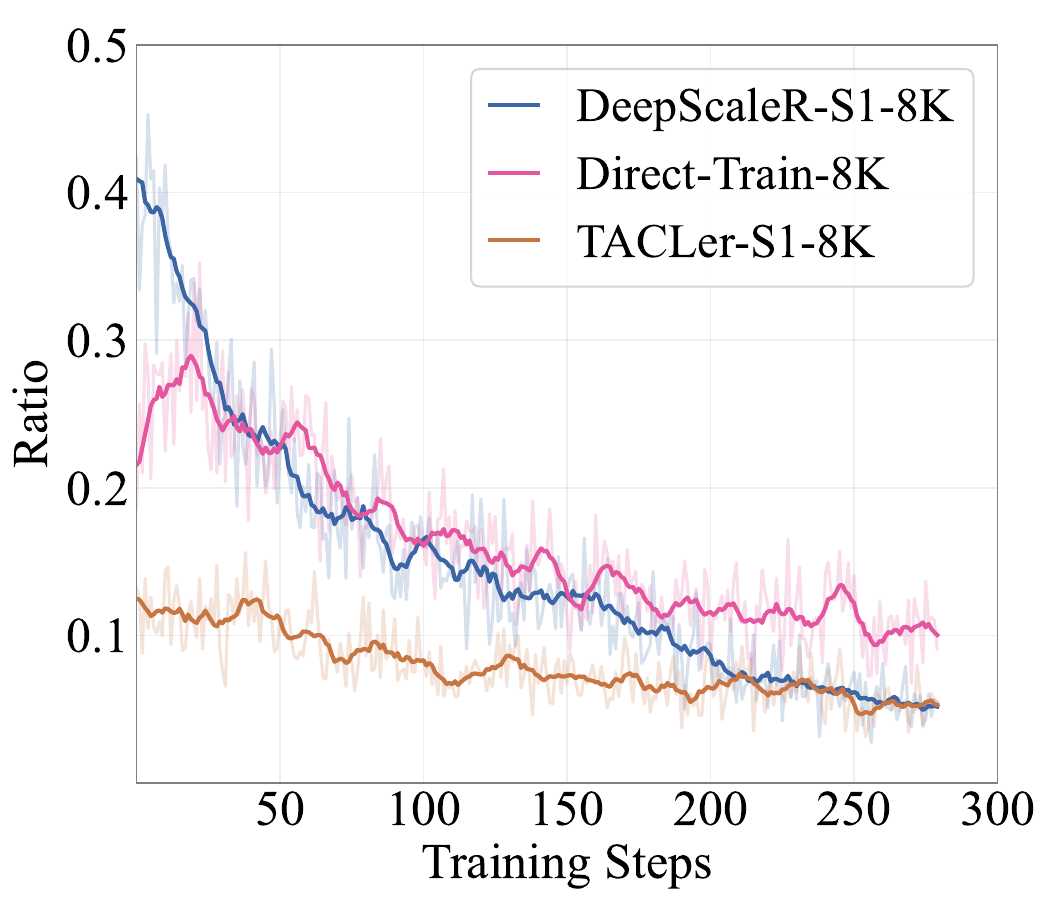}
        \caption{Response clip ratio.}
        \label{fig:s1-len}
    \end{subfigure}
    \begin{subfigure}[t]{0.31\linewidth}
        \centering
        \includegraphics[width=\linewidth]{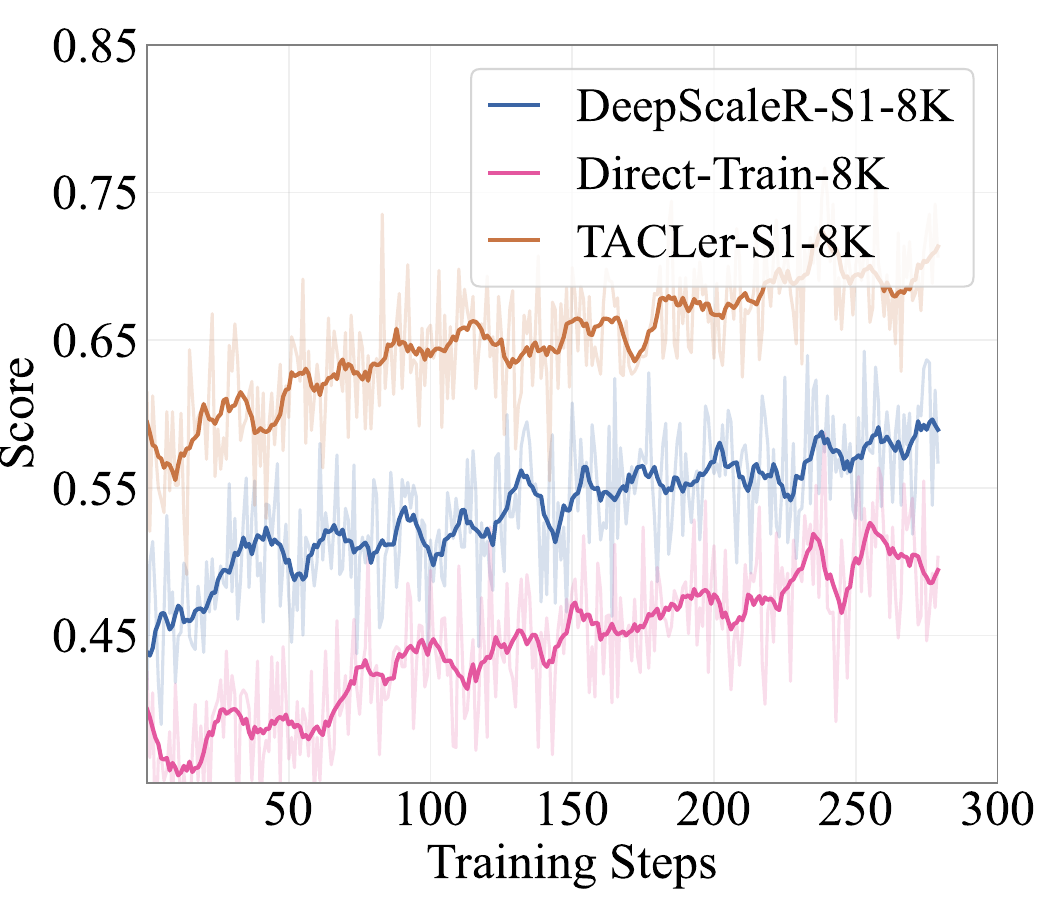}
        \caption{Response reward.}
        \label{fig:s1-reward}
    \end{subfigure}
    \caption{Comparison of TACLer, Direct-Train, and DeepScaleR-1.5B-Preview in terms of response clip ratio, length, and reward during the first training stage.}
    \label{fig:s1-clip-len-reward}
    \vspace{-0.8em}
\end{figure*}

Table~\ref{tab:training-info} summarises the training configurations for different long thinking models to compare their computational requirements. Specifically, DeepScaleR uses progressively increasing context lengths from 8K to 24K and rollout numbers from 8 to 24 in three training stages, while FastCuRL employs a similar strategy with an additional stage. In our settings, TACLer maintains a consistent context length of 8K and rollout of 8 across all three stages.
Regarding computational cost, ~\citet{deepscaler2025} indicate that doubling the context window size increases training computation by at least 2 times. At the same time, in our experiments we find that doubling the number of rollouts will roughly increase the computation by about 1 time, indicating that TACLer reduces the required training compute by over 50\% compared to the baseline models. To validate learning efficiency, in Figure~\ref{fig:s1-clip-len-reward} we present the response length, clip ratio, and reward for TACLer and Direct-Train in the first training stage (280 steps), and compare them with DeepScaleR-1.5B-Preview. Compared to the other two models, TACLer generates the shortest responses, indicating faster and more efficient reasoning. This results in a lower response clipping rate, leading to more efficient training. Furthermore, TACLer has a higher reward during training. These observations highlight TACLer's efficient learning abilities.

\begin{table}[!t]
\caption{\label{tab:data-complexity}
Data complexity statistics for three training stages, as assessed by R1-Qwen (G1: instances with correct answers; G2: instances with complete responses but incorrect answers; G3: instances with incomplete responses, see Section~\ref{sub:tailored}.)
}
\centering
\footnotesize
\setlength{\tabcolsep}{3pt}
\begin{tabular}{lrrrr}
\toprule
 & \makecell[c]{G1 (\%)} & \makecell[c]{G2 (\%)} & \makecell[c]{G3 (\%)} & Total (\#)\\
\hline
Stage 1 & 78.1 & 21.9 & 0 & 18110\\
Stage 2 & 51.1 & 14.3 & 34.6 & 27692\\
Stage 3 & 35.1 & 9.8 & 55.1 & 40315\\
\bottomrule
\end{tabular}
\end{table}

\paragraph{Data Complexity in Curriculum Learning}
The model resulting after each training stage is used to curate the difficulty-based dataset to be used in the next stage(s) of tailored curriculum learning. Table~\ref{tab:data-complexity} summarises the data composition across the three iterative stages, each consisting of three sets of data of varying difficulty, based on the results of the model R1-Qwen. As training progresses, the proportion of harder data increases from 0\% in the first stage to 34.6\% in the second stage and 55.1\% in the third stage, while the proportion of easier data decreases accordingly.

\begin{figure*}[h]
    \centering
    \begin{subfigure}[t]{0.92\linewidth}
        \centering
        \includegraphics[width=\linewidth]{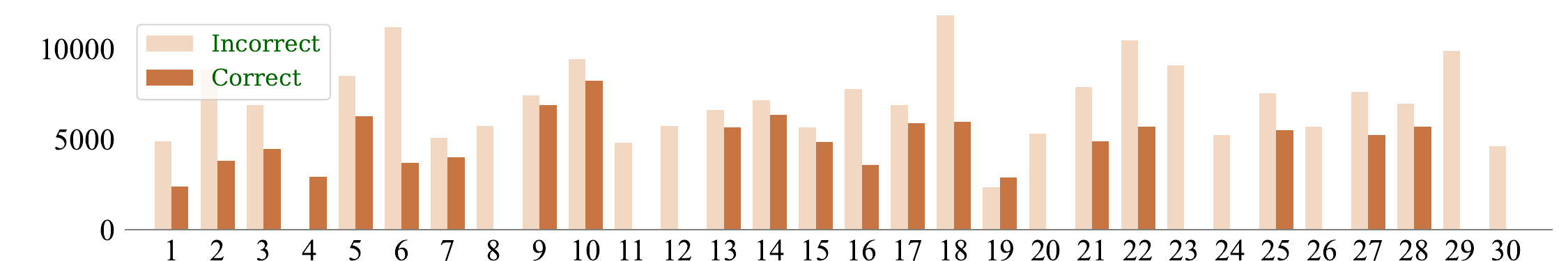}
        \caption{Average length of responses to each question under \textit{NoThinking} mode.}
        \label{fig:nonthinking}
    \end{subfigure}
    \hfill
    \begin{subfigure}[t]{0.94\linewidth}
        \centering
        \includegraphics[width=\linewidth]{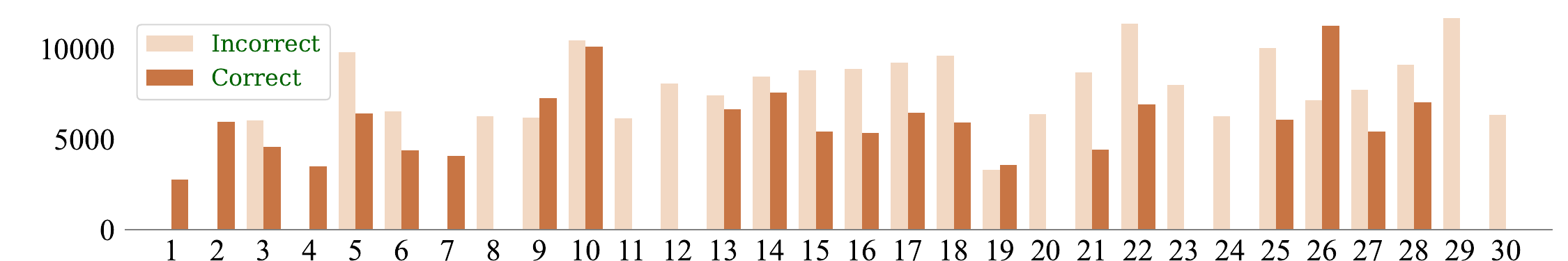}
        \caption{Average length of responses to each question under \textit{Thinking} mode.}
        \label{fig:thinking}
    \end{subfigure}
    \caption{Comparison of average response length between correct and incorrect answers. For each problem, we generate 16 responses and compute the average length separately for correct and incorrect responses; when all responses are of one type (correct or incorrect), the average length for the opposite type is set to 0.}
    \label{fig:co-in-len}
\end{figure*}

\paragraph{Contrasting Response Lengths in Correct and Incorrect Reasoning}
Figure~\ref{fig:co-in-len} presents the average response length for correct and incorrect answers at the problem level in two reasoning modes. Regarding answer correctness, we find that in both reasoning modes, correct responses are shorter than incorrect ones, as the latter often contain more verbose or repetitive content, an observation consistent with previous work~\citep{song2025fastcurl}. When comparing the responses between the two reasoning modes, correct answers in the \textit{Nothinking} mode tend to have shorter responses than those in the \textit{Thinking} mode, whereas incorrect answers show no clear length trend between the two modes.

\section{Related Work}
\paragraph{Curriculum Learning for LLM Reasoning} 
There are two main paradigms of curriculum learning: data-level, which gradually increases the complexity of training samples~\citep{bengio2009curriculum}, and model-level, which progressively increases the modelling capacity of the neural network by adding or activating more neural units as training proceeds.~\citep{karras2018progressive, morerio2017curri}. 
Recently, curriculum learning has been adopted in LLM reasoning through two main strategies: (i) progressively scheduling training data from easy to hard using criteria such as input length~\citep{song2025fastcurl}, human annotation~\citep{parashar2025cur}, and model assessments~\citep{parashar2025cur, ji2025difficulty}; and (ii) designing curriculum rewards for RL training~\citep{chen2025self}. 
In this work, 
A \textbf{key innovation} of our work relative to previous work is the introduction of a \textbf{model-tailored curriculum learning} framework which does not take an absolute concept of difficulty. Rather, \textbf{it determines what knowledge a model lacks and
needs to learn in progressive stages, thereby facilitating more efficient and effective training}.\footnote{A direct comparison with the models proposed in \citep{parashar2025cur, ji2025difficulty, chen2025self} is not possible due to the lack of publicly available checkpoints; therefore, these models are not included in Tables~\ref{tab:thinking} and~\ref{tab:nothinking}.}


\paragraph{Reinforcement Learning for LLM Reasoning}
Language models have shown impressive abilities in addressing complex reasoning tasks, leveraging CoT reasoning~\citep{wei2022chain}, reasoning consistency~\citep{wang2023selfconsistency, zhou-etal-2024-paraphrase, lai2025multi} and multilingual reasoning~\citep{shi2023language, lai-nissim-2024-mcot, chai2025xcot}. 
To further enhance reasoning capabilities, RL has emerged as an effective paradigm for training LLMs. Recent works such as OpenAI o1~\citep{openai2025o1} and DeepSeek R1~\citep{deepseekai2025r1} show that large-scale RL training can effectively elicit long CoT, leading to substantial gains on challenging mathematical reasoning and coding benchmarks. Long CoT reasoning involves a detailed iterative exploration and reflection process that is performed within a given problem space by test-time scaling~\citep{chen2025survey}. However, excessively long CoT might induce ``overthinking'', resulting in overly verbose reasoning and increased computational overhead~\citep{sui2025stop}. To address this problem, a line of work has emerged on efficient reasoning strategies, such as adaptive thinking~\citep{zhang2025adaptthink, tu2025autothink}, length-based rewards~\citep{arora2025tlmre, shen2025dast}, and disabling the thinking process~\citep{ma2025reasoning}. In this work, we propose a model-tailored curriculum reinforcement learning framework, showing that the coexistence of \textit{Thinking} and \textit{NoThinking} modes can be mutually beneficial.

\section{Conclusion}
TACLer is \textbf{a novel model-tailored curriculum reinforcement learning framework for efficient reasoning}. Leveraging staggered training based on progressive data complexity, TACLer enables the language model to learn more efficiently and also effectively. Moreover, by adopting a hybrid 
reasoning mode, TACLer shows an effective balance of accuracy and efficiency. Our experiments on the math reasoning task with four benchmarks show that \textbf{TACLer reduces training compute by over 50\% and inference token usage by over 42\% while improving accuracy by over 9\%}.  A future direction is to further extend our framework to other reasoning tasks or even to general-purpose scenarios. A clear limitation of our method, however, is the additional compute required during inference at each iteration, though the overhead remains relatively small.

\section*{Impact Statement}

This work investigates the learning and reasoning efficiency of language models on complex problem-solving tasks. By introducing a model-tailored curriculum reinforcement learning framework with a hybrid \textit{Thinking}/\textit{NoThinking} paradigm, our approach substantially reduces computational costs while improving reasoning efficiency and accuracy. This makes advanced reasoning models more accessible to people with limited resources, and supports more sustainable AI development. Beyond mathematical reasoning, our framework is potentially applicable to a wide range of domains that require multi-step reasoning. We emphasise the ethical and responsible use of our methods, aiming to lay a solid foundation for robust reasoning models and to contribute to the advancement of secure and reliable AI systems.


\bibliography{example_paper}

\begin{thebibliography}{39}
\providecommand{\natexlab}[1]{#1}
\providecommand{\url}[1]{\texttt{#1}}
\expandafter\ifx\csname urlstyle\endcsname\relax
  \providecommand{\doi}[1]{doi: #1}\else
  \providecommand{\doi}{doi: \begingroup \urlstyle{rm}\Url}\fi

\bibitem[Arora \& Zanette(2025)Arora and Zanette]{arora2025tlmre}
Arora, D. and Zanette, A.
\newblock Training language models to reason efficiently.
\newblock \emph{arXiv preprint arXiv:2502.04463}, 2025.
\newblock URL \url{https://arxiv.org/abs/2502.04463}.

\bibitem[Bengio et~al.(2009)Bengio, Louradour, Collobert, and Weston]{bengio2009curriculum}
Bengio, Y., Louradour, J., Collobert, R., and Weston, J.
\newblock Curriculum learning.
\newblock In \emph{Proceedings of the 26th Annual International Conference on Machine Learning}, pp.\  41–48, 2009.
\newblock URL \url{https://doi.org/10.1145/1553374.1553380}.

\bibitem[Chai et~al.(2025)Chai, Yang, Sun, Guo, Liu, Wang, Liang, Bai, Li, Peng, and Li]{chai2025xcot}
Chai, L., Yang, J., Sun, T., Guo, H., Liu, J., Wang, B., Liang, X., Bai, J., Li, T., Peng, Q., and Li, Z.
\newblock xcot: cross-lingual instruction timing for cross-lingual chain-of-thought reasoning.
\newblock In \emph{Proceedings of the Thirty-Ninth AAAI Conference on Artificial Intelligence and Thirty-Seventh Conference on Innovative Applications of Artificial Intelligence and Fifteenth Symposium on Educational Advances in Artificial Intelligence}, 2025.
\newblock URL \url{https://doi.org/10.1609/aaai.v39i22.34524}.

\bibitem[Chen et~al.(2025{\natexlab{a}})Chen, Qin, Liu, Peng, Guan, Wang, Hu, Zhou, Gao, and Che]{chen2025survey}
Chen, Q., Qin, L., Liu, J., Peng, D., Guan, J., Wang, P., Hu, M., Zhou, Y., Gao, T., and Che, W.
\newblock Towards reasoning era: A survey of long chain-of-thought for reasoning large language models.
\newblock \emph{arXiv preprint arXiv:2503.09567}, 2025{\natexlab{a}}.
\newblock URL \url{https://arxiv.org/abs/2503.09567}.

\bibitem[Chen et~al.(2025{\natexlab{b}})Chen, Lu, Kim, Zhang, Tang, Piché, Gontier, Bengio, and Kamalloo]{chen2025self}
Chen, X., Lu, J., Kim, M., Zhang, D., Tang, J., Piché, A., Gontier, N., Bengio, Y., and Kamalloo, E.
\newblock Self-evolving curriculum for llm reasoning.
\newblock \emph{arXiv preprint arXiv:2505.14970}, 2025{\natexlab{b}}.
\newblock URL \url{https://arxiv.org/abs/2505.14970}.

\bibitem[Chen et~al.(2025{\natexlab{c}})Chen, Xu, Liang, He, Pang, Yu, Song, Liu, Zhou, Zhang, Wang, Tu, Mi, and Yu]{chen2025do}
Chen, X., Xu, J., Liang, T., He, Z., Pang, J., Yu, D., Song, L., Liu, Q., Zhou, M., Zhang, Z., Wang, R., Tu, Z., Mi, H., and Yu, D.
\newblock Do {NOT} think that much for 2+3=? on the overthinking of long reasoning models.
\newblock In \emph{Forty-second International Conference on Machine Learning}, 2025{\natexlab{c}}.
\newblock URL \url{https://openreview.net/forum?id=MSbU3L7V00}.

\bibitem[Chen et~al.(2025{\natexlab{d}})Chen, Min, Zhang, Chen, Jiang, Cheng, Zhao, Liu, Miao, Lu, Fang, Wang, and Wen]{chen2025an}
Chen, Z., Min, Y., Zhang, B., Chen, J., Jiang, J., Cheng, D., Zhao, W.~X., Liu, Z., Miao, X., Lu, Y., Fang, L., Wang, Z., and Wen, J.-R.
\newblock An empirical study on eliciting and improving r1-like reasoning models.
\newblock \emph{arXiv preprint arXiv:2503.04548}, 2025{\natexlab{d}}.
\newblock URL \url{https://arxiv.org/pdf/2503.04548}.

\bibitem[Guo et~al.(2025)Guo, Yang, Zhang, and et~al.]{deepseekai2025r1}
Guo, D., Yang, D., Zhang, H., and et~al.
\newblock Deepseek-r1 incentivizes reasoning in llms through reinforcement learning.
\newblock \emph{Nature}, pp.\  633–638, 2025.
\newblock URL \url{https://doi.org/10.1038/s41586-025-09422-z}.

\bibitem[Hu et~al.(2025)Hu, Zhang, Han, Jiang, Zhang, and Shum]{hu2025open}
Hu, J., Zhang, Y., Han, Q., Jiang, D., Zhang, X., and Shum, H.-Y.
\newblock Open-reasoner-zero: An open source approach to scaling up reinforcement learning on the base model.
\newblock \emph{arXiv preprint arXiv:2503.24290}, 2025.
\newblock URL \url{https://arxiv.org/abs/2503.24290}.

\bibitem[Ji et~al.(2025)Ji, Zhao, Tian, Wang, Chen, Peng, Zhao, and Li]{ji2025difficulty}
Ji, Y., Zhao, S., Tian, X., Wang, H., Chen, S., Peng, Y., Zhao, H., and Li, X.
\newblock How difficulty-aware staged reinforcement learning enhances llms' reasoning capabilities: A preliminary experimental study.
\newblock \emph{arXiv preprint arXiv:2504.00829}, 2025.
\newblock URL \url{https://arxiv.org/abs/2504.00829}.

\bibitem[Karras et~al.(2018)Karras, Aila, Laine, and Lehtinen]{karras2018progressive}
Karras, T., Aila, T., Laine, S., and Lehtinen, J.
\newblock Progressive growing of {GAN}s for improved quality, stability, and variation.
\newblock In \emph{International Conference on Learning Representations}, 2018.
\newblock URL \url{https://openreview.net/forum?id=Hk99zCeAb}.

\bibitem[Kwon et~al.(2023)Kwon, Li, Zhuang, Sheng, Zheng, Yu, Gonzalez, Zhang, and Stoica]{woosuk2023efficient}
Kwon, W., Li, Z., Zhuang, S., Sheng, Y., Zheng, L., Yu, C.~H., Gonzalez, J., Zhang, H., and Stoica, I.
\newblock Efficient memory management for large language model serving with pagedattention.
\newblock In \emph{Proceedings of the 29th Symposium on Operating Systems Principles}, pp.\  611–626, 2023.
\newblock URL \url{https://doi.org/10.1145/3600006.3613165}.

\bibitem[Lai \& Nissim(2024)Lai and Nissim]{lai-nissim-2024-mcot}
Lai, H. and Nissim, M.
\newblock m{C}o{T}: Multilingual instruction tuning for reasoning consistency in language models.
\newblock In \emph{Proceedings of the 62nd Annual Meeting of the Association for Computational Linguistics (Volume 1: Long Papers)}, pp.\  12012--12026, 2024.
\newblock URL \url{https://aclanthology.org/2024.acl-long.649/}.

\bibitem[Lai et~al.(2025)Lai, Zhang, and Nissim]{lai2025multi}
Lai, H., Zhang, X., and Nissim, M.
\newblock Multidimensional consistency improves reasoning in language models.
\newblock \emph{arXiv preprint arXiv:2503.02670}, 2025.
\newblock URL \url{https://arxiv.org/abs/2503.02670}.

\bibitem[Lightman et~al.(2024)Lightman, Kosaraju, Burda, Edwards, Baker, Lee, Leike, Schulman, Sutskever, and Cobbe]{lightman2024lets}
Lightman, H., Kosaraju, V., Burda, Y., Edwards, H., Baker, B., Lee, T., Leike, J., Schulman, J., Sutskever, I., and Cobbe, K.
\newblock Let's verify step by step.
\newblock In \emph{The Twelfth International Conference on Learning Representations}, 2024.
\newblock URL \url{https://openreview.net/forum?id=v8L0pN6EOi}.

\bibitem[Luo et~al.(2025{\natexlab{a}})Luo, Shen, He, Wang, Liu, Li, Tan, Cao, and Tao]{luo2025o1pruner}
Luo, H., Shen, L., He, H., Wang, Y., Liu, S., Li, W., Tan, N., Cao, X., and Tao, D.
\newblock O1-pruner: Length-harmonizing fine-tuning for o1-like reasoning pruning.
\newblock \emph{arXiv preprint arXiv:2501.12570}, 2025{\natexlab{a}}.
\newblock URL \url{https://arxiv.org/abs/2501.12570}.

\bibitem[Luo et~al.(2025{\natexlab{b}})Luo, Tan, Wong, Shi, Tang, Roongta, Cai, Luo, Li, Popa, and Stoica]{deepscaler2025}
Luo, M., Tan, S., Wong, J., Shi, X., Tang, W.~Y., Roongta, M., Cai, C., Luo, J., Li, L.~E., Popa, R.~A., and Stoica, I.
\newblock Deepscaler: Surpassing o1-preview with a 1.5b model by scaling rl, 2025{\natexlab{b}}.
\newblock Notion Blog.

\bibitem[Ma et~al.(2025)Ma, He, Snell, Griggs, Min, and Zaharia]{ma2025reasoning}
Ma, W., He, J., Snell, C., Griggs, T., Min, S., and Zaharia, M.
\newblock Reasoning models can be effective without thinking.
\newblock \emph{arXiv preprint arXiv:2504.09858}, 2025.
\newblock URL \url{https://arxiv.org/abs/2504.09858}.

\bibitem[Meng et~al.(2024)Meng, Xia, and Chen]{meng2024simpo}
Meng, Y., Xia, M., and Chen, D.
\newblock Sim{PO}: Simple preference optimization with a reference-free reward.
\newblock In \emph{The Thirty-eighth Annual Conference on Neural Information Processing Systems}, 2024.
\newblock URL \url{https://openreview.net/forum?id=3Tzcot1LKb}.

\bibitem[Morerio et~al.(2017)Morerio, Cavazza, Volpi, Vidal, and Murino]{morerio2017curri}
Morerio, P., Cavazza, J., Volpi, R., Vidal, R., and Murino, V.
\newblock { Curriculum Dropout }.
\newblock In \emph{2017 IEEE International Conference on Computer Vision (ICCV)}, pp.\  3564--3572, 2017.
\newblock URL \url{https://doi.ieeecomputersociety.org/10.1109/ICCV.2017.383}.

\bibitem[Muennighoff et~al.(2025)Muennighoff, Yang, Shi, Li, Fei-Fei, Hajishirzi, Zettlemoyer, Liang, Candès, and Hashimoto]{muennighoff2025s1}
Muennighoff, N., Yang, Z., Shi, W., Li, X.~L., Fei-Fei, L., Hajishirzi, H., Zettlemoyer, L., Liang, P., Candès, E., and Hashimoto, T.
\newblock s1: Simple test-time scaling.
\newblock \emph{arXiv preprint arXiv:2501.19393}, 2025.
\newblock URL \url{https://arxiv.org/abs/2501.19393}.

\bibitem[OpenAI(2025)]{openai2025o1}
OpenAI.
\newblock Competitive programming with large reasoning models.
\newblock \emph{arXiv preprint arXiv:2502.06807}, 2025.
\newblock URL \url{https://arxiv.org/abs/2502.06807}.

\bibitem[Parashar et~al.(2025)Parashar, Gui, Li, Ling, Vemuri, Olson, Li, Zhang, Caverlee, Kalathil, and Ji]{parashar2025cur}
Parashar, S., Gui, S., Li, X., Ling, H., Vemuri, S., Olson, B., Li, E., Zhang, Y., Caverlee, J., Kalathil, D., and Ji, S.
\newblock Curriculum reinforcement learning from easy to hard tasks improves llm reasoning.
\newblock \emph{arXiv preprint arXiv:2506.06632}, 2025.
\newblock URL \url{https://arxiv.org/abs/2506.06632}.

\bibitem[Shao et~al.(2024)Shao, Wang, Zhu, Xu, Song, Bi, Zhang, Zhang, Li, Wu, and Guo]{shao2024deepseek}
Shao, Z., Wang, P., Zhu, Q., Xu, R., Song, J., Bi, X., Zhang, H., Zhang, M., Li, Y.~K., Wu, Y., and Guo, D.
\newblock Deepseekmath: Pushing the limits of mathematical reasoning in open language models.
\newblock \emph{arXiv preprint arXiv:2402.03300}, 2024.
\newblock URL \url{https://arxiv.org/abs/2402.03300}.

\bibitem[Shen et~al.(2025)Shen, Zhang, Huang, Shi, Zhang, Yan, Wang, Wang, Liu, and Lian]{shen2025dast}
Shen, Y., Zhang, J., Huang, J., Shi, S., Zhang, W., Yan, J., Wang, N., Wang, K., Liu, Z., and Lian, S.
\newblock Dast: Difficulty-adaptive slow-thinking for large reasoning models.
\newblock \emph{arXiv preprint arXiv:2503.04472}, 2025.
\newblock URL \url{https://arxiv.org/abs/2503.04472}.

\bibitem[Sheng et~al.(2025)Sheng, Zhang, Ye, Wu, Zhang, Zhang, Peng, Lin, and Wu]{sheng2025hybrid}
Sheng, G., Zhang, C., Ye, Z., Wu, X., Zhang, W., Zhang, R., Peng, Y., Lin, H., and Wu, C.
\newblock Hybridflow: A flexible and efficient rlhf framework.
\newblock In \emph{Proceedings of the Twentieth European Conference on Computer Systems}, pp.\  1279–1297, 2025.
\newblock URL \url{https://doi.org/10.1145/3689031.3696075}.

\bibitem[Shi et~al.(2023)Shi, Suzgun, Freitag, Wang, Srivats, Vosoughi, Chung, Tay, Ruder, Zhou, Das, and Wei]{shi2023language}
Shi, F., Suzgun, M., Freitag, M., Wang, X., Srivats, S., Vosoughi, S., Chung, H.~W., Tay, Y., Ruder, S., Zhou, D., Das, D., and Wei, J.
\newblock Language models are multilingual chain-of-thought reasoners.
\newblock In \emph{The Eleventh International Conference on Learning Representations}, 2023.
\newblock URL \url{https://openreview.net/forum?id=fR3wGCk-IXp}.

\bibitem[Song et~al.(2025)Song, Zheng, Li, Yang, Luo, Pan, and Zhang]{song2025fastcurl}
Song, M., Zheng, M., Li, Z., Yang, W., Luo, X., Pan, Y., and Zhang, F.
\newblock Fastcurl: Curriculum reinforcement learning with progressive context extension for efficient training r1-like reasoning models.
\newblock \emph{arXiv preprint arXiv:2503.17287}, 2025.
\newblock URL \url{https://arxiv.org/abs/2503.17287}.

\bibitem[Sui et~al.(2025)Sui, Chuang, Wang, Zhang, Zhang, Yuan, Liu, Wen, Zhong, Chen, and Hu]{sui2025stop}
Sui, Y., Chuang, Y.-N., Wang, G., Zhang, J., Zhang, T., Yuan, J., Liu, H., Wen, A., Zhong, S., Chen, H., and Hu, X.
\newblock Stop overthinking: A survey on efficient reasoning for large language models.
\newblock \emph{arXiv preprint arXiv:2503.16419}, 2025.
\newblock URL \url{https://arxiv.org/abs/2503.16419}.

\bibitem[Tu et~al.(2025)Tu, Lin, Zhang, Tian, Li, Lan, and Zhao]{tu2025autothink}
Tu, S., Lin, J., Zhang, Q., Tian, X., Li, L., Lan, X., and Zhao, D.
\newblock Learning when to think: Shaping adaptive reasoning in r1-style models via multi-stage rl.
\newblock \emph{arXiv preprint arXiv:2505.10832}, 2025.
\newblock URL \url{https://arxiv.org/abs/2505.10832}.

\bibitem[Wang et~al.(2023)Wang, Wei, Schuurmans, Le, Chi, Narang, Chowdhery, and Zhou]{wang2023selfconsistency}
Wang, X., Wei, J., Schuurmans, D., Le, Q.~V., Chi, E.~H., Narang, S., Chowdhery, A., and Zhou, D.
\newblock Self-consistency improves chain of thought reasoning in language models.
\newblock In \emph{The Eleventh International Conference on Learning Representations}, 2023.
\newblock URL \url{https://openreview.net/forum?id=1PL1NIMMrw}.

\bibitem[Wei et~al.(2022)Wei, Wang, Schuurmans, Bosma, brian ichter, Xia, Chi, Le, and Zhou]{wei2022chain}
Wei, J., Wang, X., Schuurmans, D., Bosma, M., brian ichter, Xia, F., Chi, E.~H., Le, Q.~V., and Zhou, D.
\newblock Chain of thought prompting elicits reasoning in large language models.
\newblock In \emph{Advances in Neural Information Processing Systems}, 2022.
\newblock URL \url{https://openreview.net/forum?id=_VjQlMeSB_J}.

\bibitem[Wu et~al.(2025)Wu, Yao, Liu, Liu, Fu, Han, Li, Zhen, Zhong, and Yuan]{wu2025unlock}
Wu, H., Yao, Y., Liu, S., Liu, Z., Fu, X., Han, X., Li, X., Zhen, H.-L., Zhong, T., and Yuan, M.
\newblock Unlocking efficient long-to-short llm reasoning with model merging.
\newblock \emph{arXiv preprint arXiv:2503.20641}, 2025.
\newblock URL \url{https://arxiv.org/abs/2503.20641}.

\bibitem[Wu et~al.(2024)Wu, Sun, Li, Welleck, and Yang]{wu2024scaling}
Wu, Y., Sun, Z., Li, S., Welleck, S., and Yang, Y.
\newblock Inference scaling laws: An empirical analysis of compute-optimal inference for llm problem-solving.
\newblock In \emph{The 4th Workshop on Mathematical Reasoning and AI at NeurIPS'24}, 2024.
\newblock URL \url{https://openreview.net/forum?id=j7DZWSc8qu}.

\bibitem[Yu et~al.(2025)Yu, Zhang, Zhu, Yuan, Zuo, Yue, Dai, Fan, Liu, Liu, Liu, Lin, Lin, Ma, Sheng, Tong, Zhang, Zhang, Zhang, Zhu, Zhu, Chen, Chen, Wang, Yu, Song, Wei, Zhou, Liu, Ma, Zhang, Yan, Qiao, Wu, and Wang]{yu2025dapo}
Yu, Q., Zhang, Z., Zhu, R., Yuan, Y., Zuo, X., Yue, Y., Dai, W., Fan, T., Liu, G., Liu, L., Liu, X., Lin, H., Lin, Z., Ma, B., Sheng, G., Tong, Y., Zhang, C., Zhang, M., Zhang, W., Zhu, H., Zhu, J., Chen, J., Chen, J., Wang, C., Yu, H., Song, Y., Wei, X., Zhou, H., Liu, J., Ma, W.-Y., Zhang, Y.-Q., Yan, L., Qiao, M., Wu, Y., and Wang, M.
\newblock Dapo: An open-source llm reinforcement learning system at scale.
\newblock \emph{arXiv preprint arXiv:2503.14476}, 2025.
\newblock URL \url{https://arxiv.org/abs/2503.14476}.

\bibitem[Yue et~al.(2024)Yue, Qu, Zhang, Fu, Huang, Sun, Su, and Chen]{yue2024mammoth}
Yue, X., Qu, X., Zhang, G., Fu, Y., Huang, W., Sun, H., Su, Y., and Chen, W.
\newblock {MA}mmo{TH}: Building math generalist models through hybrid instruction tuning.
\newblock In \emph{The Twelfth International Conference on Learning Representations}, 2024.
\newblock URL \url{https://openreview.net/forum?id=yLClGs770I}.

\bibitem[Zhang et~al.(2025)Zhang, Lin, Hou, Feng, and Li]{zhang2025adaptthink}
Zhang, J., Lin, N., Hou, L., Feng, L., and Li, J.
\newblock Adaptthink: Reasoning models can learn when to think.
\newblock \emph{arXiv preprint arXiv:2505.13417}, 2025.
\newblock URL \url{https://arxiv.org/abs/2505.13417}.

\bibitem[Zhou et~al.(2023)Zhou, Sch{\"a}rli, Hou, Wei, Scales, Wang, Schuurmans, Cui, Bousquet, Le, and Chi]{zhou2023leasttomost}
Zhou, D., Sch{\"a}rli, N., Hou, L., Wei, J., Scales, N., Wang, X., Schuurmans, D., Cui, C., Bousquet, O., Le, Q.~V., and Chi, E.~H.
\newblock Least-to-most prompting enables complex reasoning in large language models.
\newblock In \emph{The Eleventh International Conference on Learning Representations}, 2023.
\newblock URL \url{https://openreview.net/forum?id=WZH7099tgfM}.

\bibitem[Zhou et~al.(2024)Zhou, Zhu, Antognini, Kim, and Zhang]{zhou-etal-2024-paraphrase}
Zhou, Y., Zhu, Y., Antognini, D., Kim, Y., and Zhang, Y.
\newblock Paraphrase and solve: Exploring and exploiting the impact of surface form on mathematical reasoning in large language models.
\newblock In \emph{Proceedings of the 2024 Conference of the North American Chapter of the Association for Computational Linguistics: Human Language Technologies (Volume 1: Long Papers)}, pp.\  2793--2804, 2024.
\newblock URL \url{https://aclanthology.org/2024.naacl-long.153/}.

\end{thebibliography}
\bibliographystyle{icml2026}

\newpage
\appendix
\onecolumn

\section{Appendix}

\subsection{Training and Evaluation Setting}
\label{app:setting}
We train TACLer based on VeRL~\citep{sheng2025hybrid} and perform inference using vLLM~\citep{woosuk2023efficient}. Across three training stages, we set the context length to 8K, the batch size to 128, and the learning rate to 1e-6.  The hyperparameters $\epsilon_{\text{low}}$ and $\epsilon_{\text{high}}$ are set to 0.2 and 0.28, respectively. TACLer is trained on 4 NVIDIA H100 GPUs (94 GB memory each) for approximately 96 hours. Regarding the cost of curriculum labelling during inference, processing the full dataset takes approximately 5 hours on a single H100 GPU. After the first and second training stages, this time is further reduced due to the substantial improvement in TACLer’s reasoning efficiency. Overall, the incurred overhead is relatively small.

We report both the average reasoning accuracy and the response length based on the number of tokens in our evaluation. Following previous work~\citep{zhang2025adaptthink} we adopt the sampling method with a context size of 16K, a temperature of 0.6 and a top-$p$ value of 0.95 to generate the response for each question. We sample 16 responses for each question in AIME 2024 and AIME 2025, and report the average results due to their limited dataset size.

\subsection{Baselines}
\label{app:baselines}

We compare TACLer with state-of-the-art 1.5B baselines on efficient math reasoning.

\begin{itemize}
\item \textbf{OverThink}~\citep{chen2025do}: a SimPO~\citep{meng2024simpo} based method, taking the longest response as the negative example and the first two correct reasoning as the positive ones. 
\item \textbf{DAST}~\citep{shen2025dast}: a SimPO-based method, where preference data is ranked by pre-sampled responses using a length-based reward function. 
\item \textbf{O1-Pruner}~\citep{luo2025o1pruner}: it builds a baseline via pre-sampling and guides the model to generate more concise reasoning under accuracy constraints. 
\item \textbf{TLMRE}~\citep{arora2025tlmre}: a method that incorporates a length-based penalty in on-policy RL training. 
\item \textbf{ModelMerging}~\citep{wu2025unlock}: it takes a weighted average of the weights of the reasoning model and the non-reaosning model.
\item \textbf{AdaptThink}~\citep{zhang2025adaptthink}: a method trained with a constrained objective to encourage the model to reason in the NoThinking way. 
\item \textbf{AutoThink}~\citep{tu2025autothink}: an RL-based approach optimizes models to invoke long inferences only when necessary. 

\end{itemize}

All above methods use DeepScaleR dataset, where methods (1)-(5) are reproduced by ~\citet{zhang2025adaptthink}, thereby having a fair comparison. We also compare our method with several 1.5B baselines trained for long CoT reasoning in the \textit{Thinking} mode.

\begin{itemize}
\item \textbf{STILL-3}~\citep{chen2025an}: a method to enhance the model's slow thinking through RL training.

\item \textbf{DeepScaleR}~\citep{deepscaler2025}: an iterative training method that iteratively increases the context length from 8K to 24K, allowing the model to improve the quality of long CoT.

\item \textbf{FastCuRL}~\citep{song2025fastcurl}: a curriculum RL framework with progressively extended context, in which the training data complexity is categorised based on input length.

\end{itemize}

\subsection{Case Study}
\label{app:case}
\begin{figure}[t]
  \centering
\includegraphics[width=0.8\textwidth]{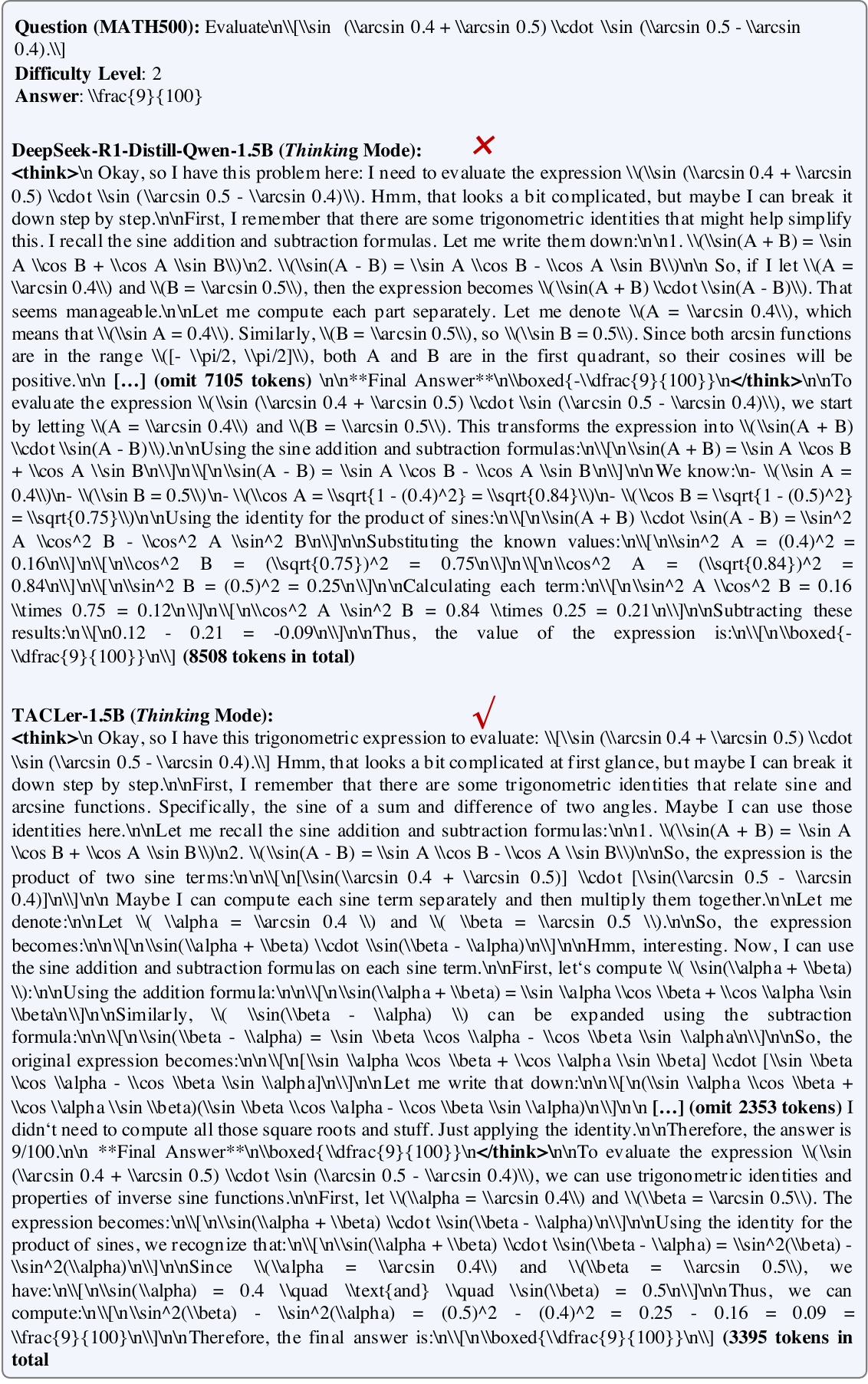}
  \caption{An example MATH500 problem at level 2 and the responses generated by R1-Qwen and TACLer under \textit{Thinking} mode.}
  \label{fig:case-ds-tacler}
\end{figure}

Figure~\ref{fig:case-ds-tacler} presents a level-2 problem from the MATH500 dataset, which is considered relatively easy by humans but proves more challenging for language models. The response generated by the base model R1-Qwen under the \textit{Thinking} mode produces an incorrect answer, while TACLer answers correctly with fewer tokens. 

\begin{figure}[t]
  \centering
\includegraphics[width=0.8\textwidth]{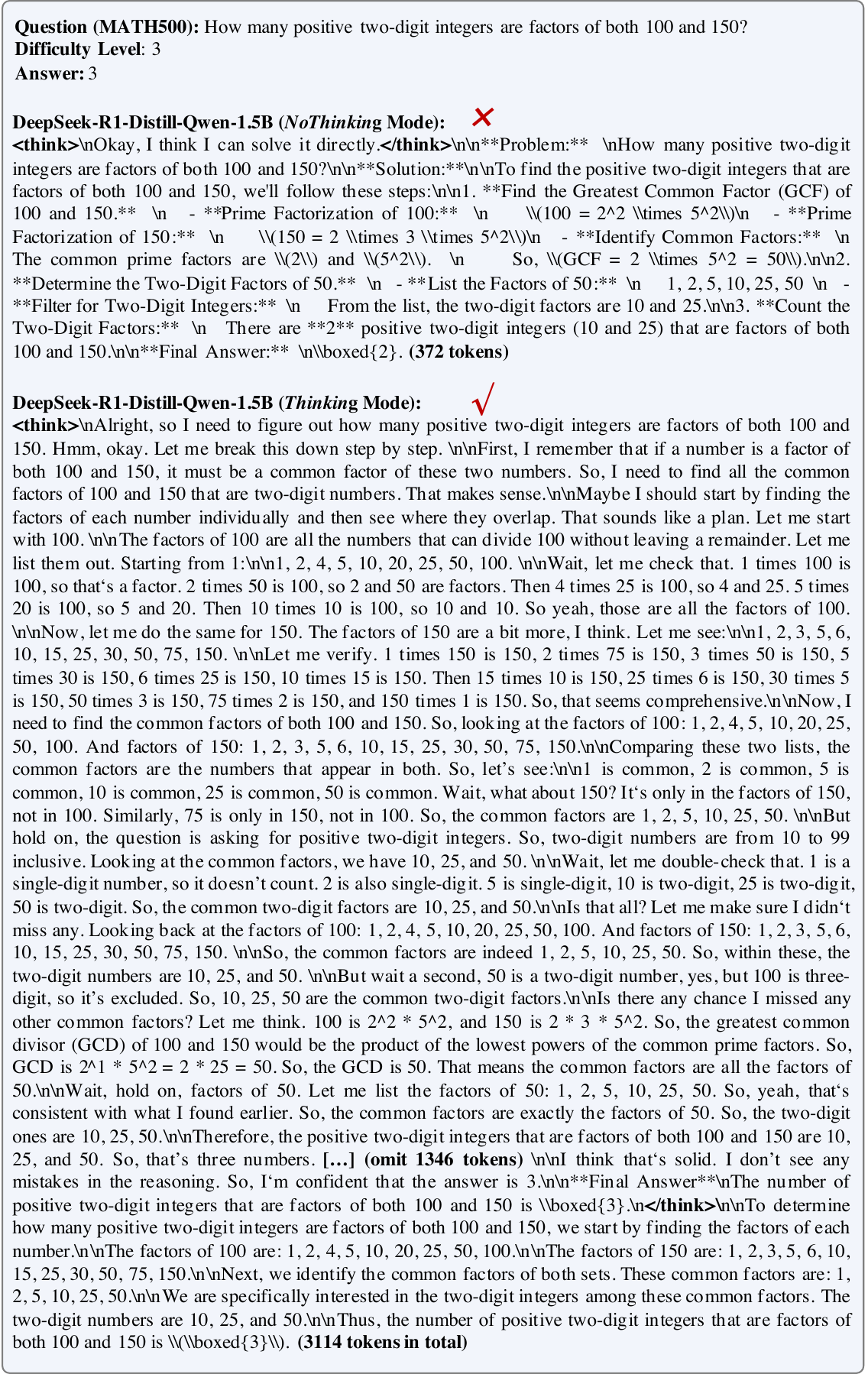}
  \caption{An example MATH500 problem and the corresponding responses generated by R1-Qwen under two reasoning modes.}
  \label{fig:case-deepseek}
\end{figure}

\begin{figure}[t]
  \centering
\includegraphics[width=0.8\textwidth]{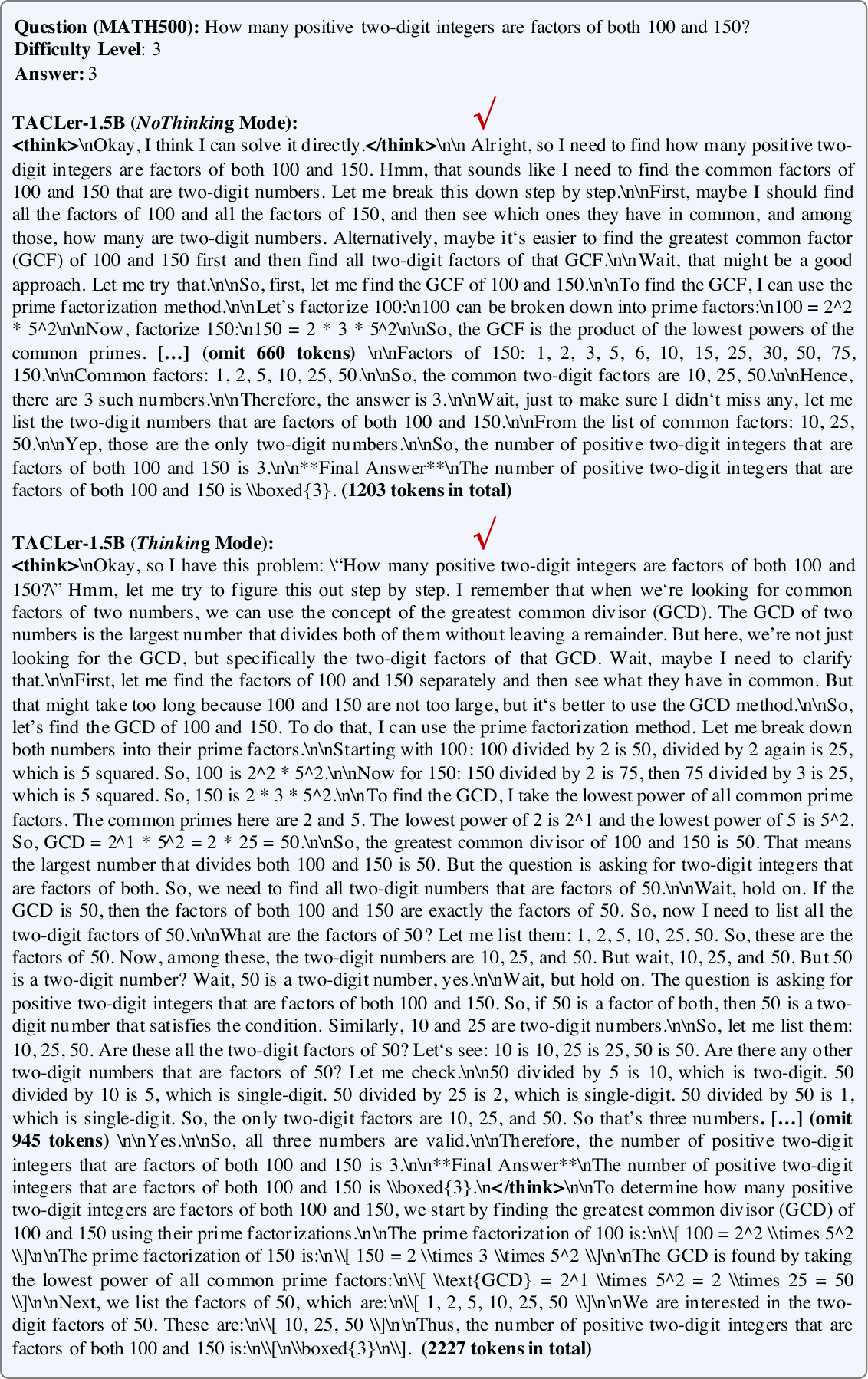}
  \caption{An example MATH500 problem and the corresponding responses generated by TACLer under two reasoning modes.}
  \label{fig:case-tacler}
\end{figure}

Figures~\ref{fig:case-deepseek} and \ref{fig:case-tacler} present a math problem at difficulty level 3, comparing outputs of the base model R1-Qwen and our model TACLer under two reasoning modes: \textit{Thinking} and  \textit{NoThinking}. For R1-Qwen, the output under the \textit{NoThinking} mode is short but the final answer is incorrect, whereas the \textit{Thinking} mode produces a correct answer at the cost of excessively long outputs. In contrast, TACLer produces shorter outputs with correct final answers in both modes, with the output under the \textit{NoThinking} mode being about half the length of that under the \textit{Thinking} mode. 
We attribute this behavior to the following reasons: (i) for \textit{NoThinking} mode, TACLer compensates for missing explicit thinking by expanding its reasoning steps, thus resulting in a slight increase in output length; (ii) for \textit{Thinking} mode, TACLer leverage a compression effect that refines its reasoning into more concise responses.
Overall, these cases highlight distinct differences in reasoning behaviour and performance between the two models across different reasoning modes.

\end{document}